\documentclass[11pt]{article}

\usepackage[preprint]{acl}

\usepackage{times}
\usepackage{latexsym}

\usepackage[T1]{fontenc}

\usepackage[utf8]{inputenc}

\usepackage{microtype}

\usepackage{inconsolata}
\usepackage{amssymb}
\usepackage{graphicx}
\usepackage{amsmath}
\usepackage{booktabs}
\usepackage{multirow}
\usepackage{tabularx}
\usepackage{longtable}
\usepackage{algorithm}
\usepackage{algpseudocode}
\usepackage{fontawesome5}

\definecolor{myred}{RGB}{156, 39, 33}
\definecolor{myblue}{RGB}{31, 90, 153}
\usepackage[table]{xcolor}

\newcommand{\poschg}[1]{\textcolor{myblue}{\scriptsize{(+#1)}}}
\newcommand{\negchg}[1]{\textcolor{myred}{\scriptsize{(-#1)}}}

\usepackage{booktabs}
\usepackage{xcolor}
\usepackage{soul}
\usepackage{array}

\title{Beyond Simply Environment Scaling: Designing Effective Environment Distributions for Multimodal Agent Learning}

\author{Kejian Zhu\textsuperscript{1,2}, Zhuoran Jin\textsuperscript{1,2}, Dongqi Huang\textsuperscript{1,2}, Hongbang Yuan\textsuperscript{1,2}, Yupu Hao\textsuperscript{1,2}, \\ \textbf{Kang Liu\textsuperscript{1,2}}, \textbf{Jun Zhao\textsuperscript{1,2}\footnotemark[2]} \\ \textsuperscript{1}The Key Laboratory of Cognition and Decision Intelligence for Complex Systems, \\ Institute of Automation, Chinese Academy of Sciences, Beijing, China \\ \textsuperscript{2}School of Artificial Intelligence, University of Chinese Academy of Sciences, Beijing, China \\
\texttt{zhukejian2025@ia.ac.cn}, \texttt{\{zhuoran.jin,kliu,jzhao\} @nlpr.ia.ac.cn } \\ 
} 

\begin{document}
\maketitle

\renewcommand{\thefootnote}{\fnsymbol{footnote}}
\footnotetext[2]{Corresponding authors.}

\begin{abstract}
Recent works train agents by constructing large-scale multimodal environment pools. However, we find that simply increasing the number of multimodal environments does not always benefit. We further analyze the limitations in current multimodal environment distributions through a series of experiments. Based on these findings, we study how to build more effective training environment distributions from two dimensions: \textbf{diversity} and \textbf{difficulty structure}. For diversity, we propose \textbf{Ability-aware Environment Selection (AES)} to obtain diverse environment sets. For difficulty structure, we propose \textbf{Hierarchical Difficulty Curriculum (HDC)}, which organizes curriculum learning through two difficulty levels: harness weakening and state-scale progression. Experiments show that AES and HDC effectively improve multimodal agent training. \href{https://github.com/GaryStack/Beyond-MMEnv-Scaling}
{\faGithub\ \textbf{Code}}

\end{abstract}

\section{Introduction}

Recent advances in multimodal large language models (MLLMs) and agents are shifting the focus of model training from static datasets~\cite{jin2026looklightthinkheavy} to dynamic environments~\cite{odysseus}. To improve different abilities of agents, a growing body of work constructs large-scale multimodal environments and trains agents jointly on them~\cite{gym-v}. Existing efforts on scaling multimodal environments mainly emphasize the quality of each individual environments~\cite{V-GameGym}. They typically verify whether an environment can be executed~\cite{AutoWebWorld} or provide reliable reward~\cite{Matrix-game}. We refer to this perspective as \textbf{sample-level quality}. Such checks are necessary for ensuring that each environment is usable. However, for agent training, the quality of individual environments is not sufficient. What an agent actually learns, and how well it generalizes, is also directly depended on the effectiveness of training environment \textbf{distribution}.

\begin{figure}[t]
\centering
  \includegraphics[width=\columnwidth]{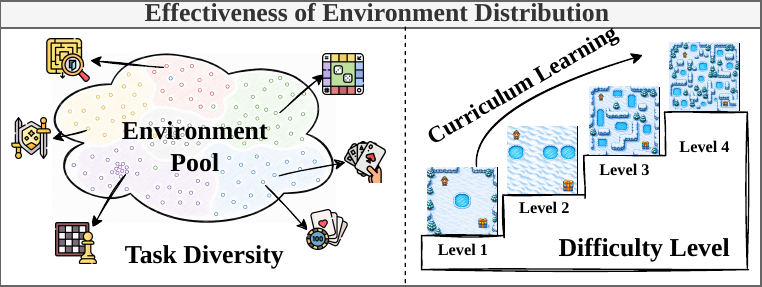}
  \caption{Environment distribution needs to be evaluated from two dimensions: diversity and difficulty level.}
  \label{fig:intro_distribution_quality}
    \vspace{-12pt}
\end{figure}



To  analyze the limitations of existing environment distributions, we build a pool of 200 environments based on prior works~\cite{visgym,gym-v} and conduct a series of preliminary experiments in Section \ref{sec:preliminaries}. First, the results in Section \ref{sec:scaling_failure} show that as the number of training environment types increases, model performance may fluctuate substantially or even degrade. This suggests that more multimodal environment types are not necessarily better. Second, in Section \ref{sec:negative_transfer}, we find that this mixed-training failure is more severe in multimodal environments. For the same environment distribution, we construct both text-symbolic and multimodal versions for comparison. The results show that the multimodal version exhibits stronger negative transfer and gradient conflicts across environments, and is more likely to suffer performance degradation under mixed training. These preliminary findings indicate that training multimodal agents requires a deeper analysis of the effectiveness of training environment distributions, going beyond simple environment scaling.



\begin{figure*}[t]
  \includegraphics[width=\linewidth]{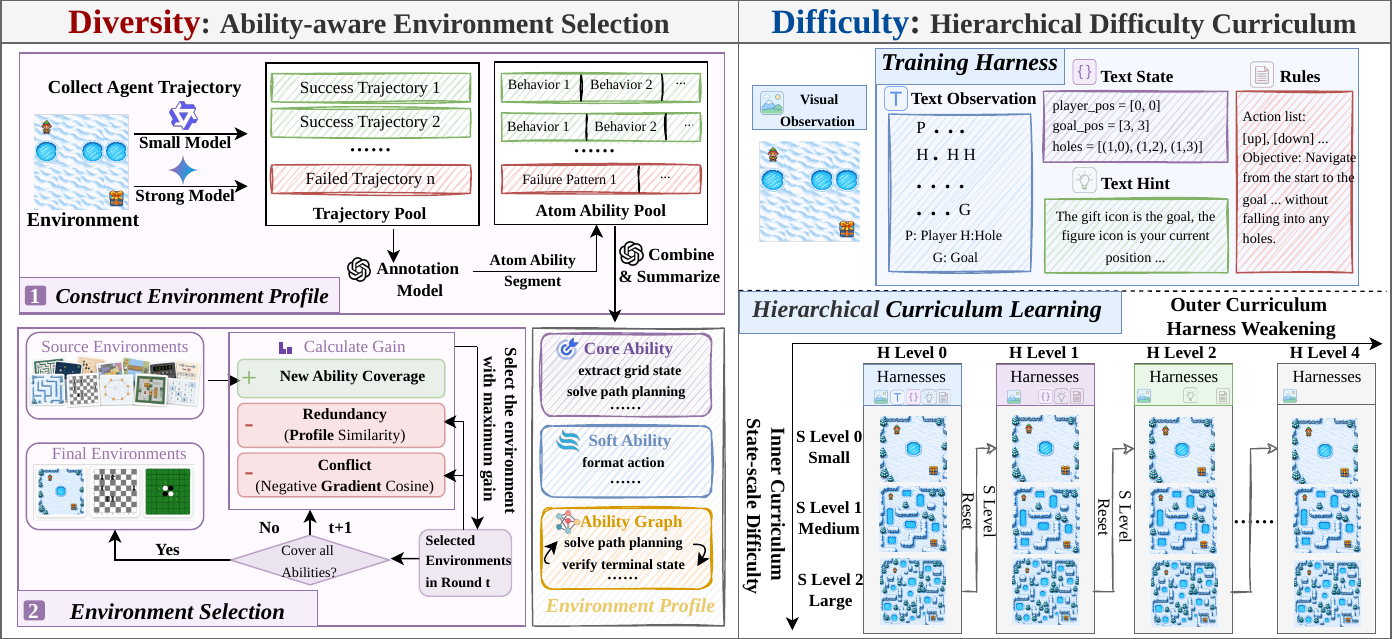}
  \caption{An overview of our methods for evaluating and designing effective environment distribution.}
  \label{fig:methodology_intro}
  \vspace{-15pt}
\end{figure*}

Thus our research question is: \textbf{how can we evaluate and design effective training environment distributions for multimodal agents?} We argue that the effectiveness of environment distribution should be evaluated along two dimensions: \textcolor{myred}{\emph{diversity}} and \textcolor{myblue}{\emph{difficulty structure}}, as shown in Figure~\ref{fig:intro_distribution_quality}. Diversity requires the environment set to cover broad abilities while avoiding redundancy and conflicts. Difficulty structure refers to whether the environment distribution provides a meaningful difficulty progression, so that the agent can improve continuously and stably. Based on this perspective, we analyze these two dimensions separately and propose corresponding methods. The overall framework of our method is illustrated in Figure~\ref{fig:methodology_intro}.

\textbf{(1) For \textcolor{myred}{diversity}}, our goal is to obtain an environment distribution with high coverage, low redundancy and optimization conflict. As discussed in Section~\ref{sec:diversity}, instead of measuring diversity through surface-level signals, such as task-description representations, as in prior works~\cite{V-GameGym}, we argue that the underlying diversity of environments should be analyzed from the perspective of agent behaviors and the abilities learned during training. We therefore analyze agent trajectories in different environments, decompose them into reusable atomic abilities, and construct a meta-ability profile for each environment. Based on these profiles, we propose \textbf{Ability-aware Environment Selection (AES)}. AES selects environments by maximizing the coverage of meta-abilities. Meanwhile, it reduces redundancy through profile similarity analysis and mitigates potential optimization conflicts through mechanistic gradient analysis.

\textbf{(2) For \textcolor{myblue}{difficulty structure}}, we further study multimodal-specific difficulty structure in Section~\ref{sec:curriculum}. We first show that two major ability bottlenecks of multimodal agents are visual state extraction and world modeling, which is consistent with prior multimodal agent studies~\cite{visgym, gym-v}. However, existing works often follow the same framework as text environments and mainly define difficulty structure by state scale, such as grid size. This does not directly target the unique bottlenecks of multimodal agents. To this end, we propose \textbf{Hierarchical Difficulty Curriculum (HDC)}, a multimodal difficulty curriculum. Our two-level difficulty structure consists of harness weakening and state-scale progression. Here, a harness refers to a type of training scaffold that provides additional textual auxiliary information to help the agent extract visual states and understand world dynamics. As shown on the right side of Figure~\ref{fig:methodology_intro}, we design multiple harnesses and gradually remove these scaffolds to help the agent overcome the two major ability bottlenecks. Within each harness stage, we further apply conventional state-scale curriculum learning as the inner curriculum to improve training stability and generalization.


We evaluate our framework on multiple settings. 
The experiments show that the high-quality environment subset selected by \textbf{AES} can outperform training on the full environment pool while using fewer environments. In addition, \textbf{HDC} leads to more stable training and better performance than both direct training and conventional scale-only curriculum learning.
Together, \textbf{AES} and \textbf{HDC} achieve a 143.2\% average relative gain across different settings. 
These results highlight that designing effective environment distributions is an important and promising direction for multimodal agent research.

\section{Preliminaries} 
\label{sec:preliminaries}

We first collect and unify 200 multimodal environments based on prior works ~\cite{visgym, gym-v}. Then we conduct a series of controlled experiments on them to analyze the key problems in current multimodal environments. Detailed settings are provided in Appendix~\ref{appendix:exp_details}. 

\subsection{Naive Scaling Multimodal Environments Not Always Benefits}
\label{sec:scaling_failure}

\begin{figure}[t]
  \includegraphics[width=0.9\columnwidth]{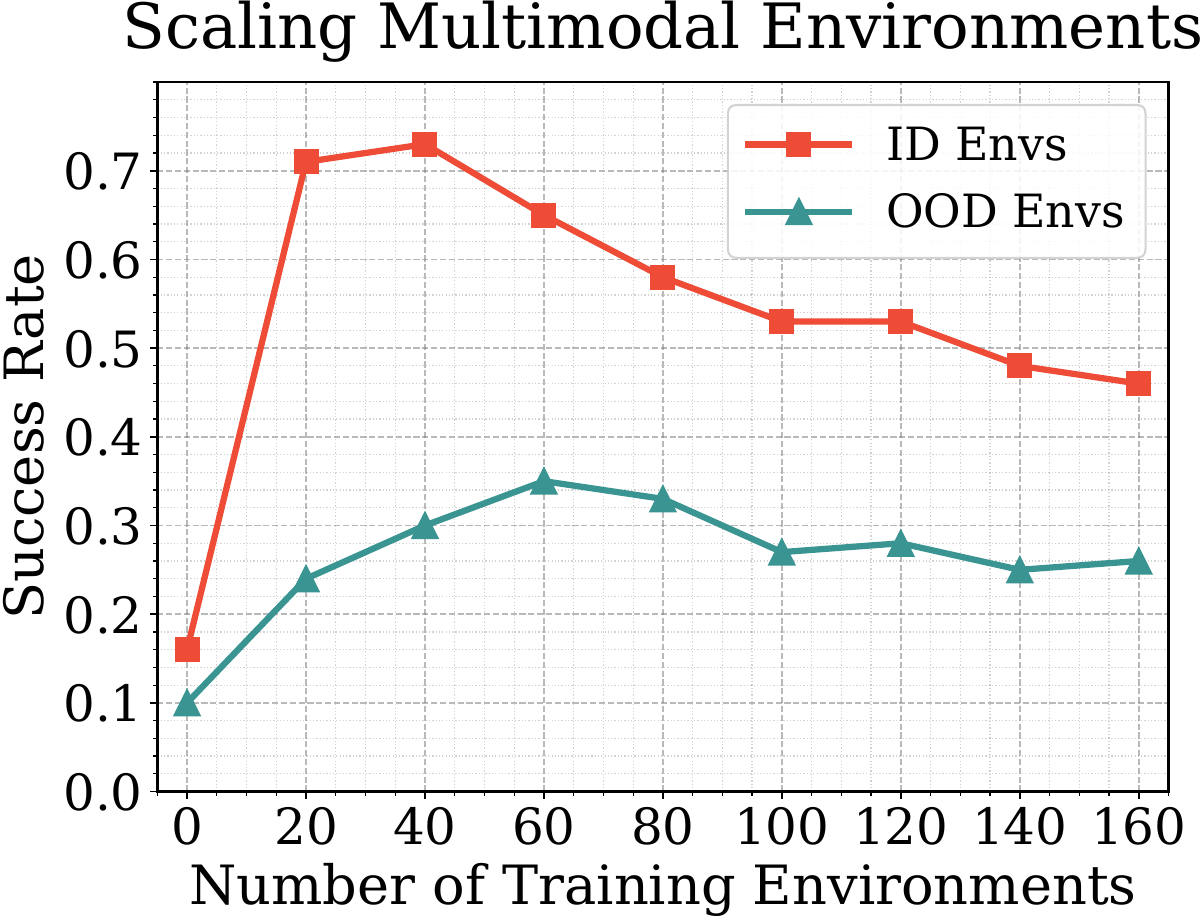}
  \caption{Effect of simply environment scaling.}
  \label{fig:preliminary_experiments_scaling}
    \vspace{-10pt}
\end{figure}

To train agents with diverse abilities, recent works have constructed large-scale multimodal environment collections. However, does simply scaling the number of environment types always benefits? To study this question, we train agents on our 200-environment pool. We gradually increase the number of environment types used for training and keep the total compute budget fixed. As shown in Figure~\ref{fig:preliminary_experiments_scaling}, model performance does not monotonically improve as the number of environment types increases. Under some environment scales, the performance may even decrease. This indicates that simply scaling the number of multimodal environments is not always stable. It may introduce redundancy or conflicts. Therefore, we need to further analyze the distribution of environment set.

\begin{table}[t]
\centering
\small
\setlength{\tabcolsep}{4.5pt}
\begin{tabular}{lccc}
\toprule
Setting & Avg. Single & Avg. Mixed & Drop \\
\midrule
Text-symbolic & 70.4 & 69.5 & -1.3\%  \\
Multimodal    & 48.4 & 43.2 & -10.7\% \\
\bottomrule
\end{tabular}
\caption{
Results of mixed-environment training.
Avg. Single denotes the average performance of separately trained single-environment models, while Avg. Mixed denotes the average performance of one model trained on the mixed environment set.
}
\label{tab:negative_transfer_summary}
    \vspace{-14pt}
\end{table}

\subsection{Multimodal Environments Are More Prone to Negative Transfer}
\label{sec:negative_transfer}

\begin{figure*}[t]
    \centering
    \includegraphics[width=0.95\linewidth]{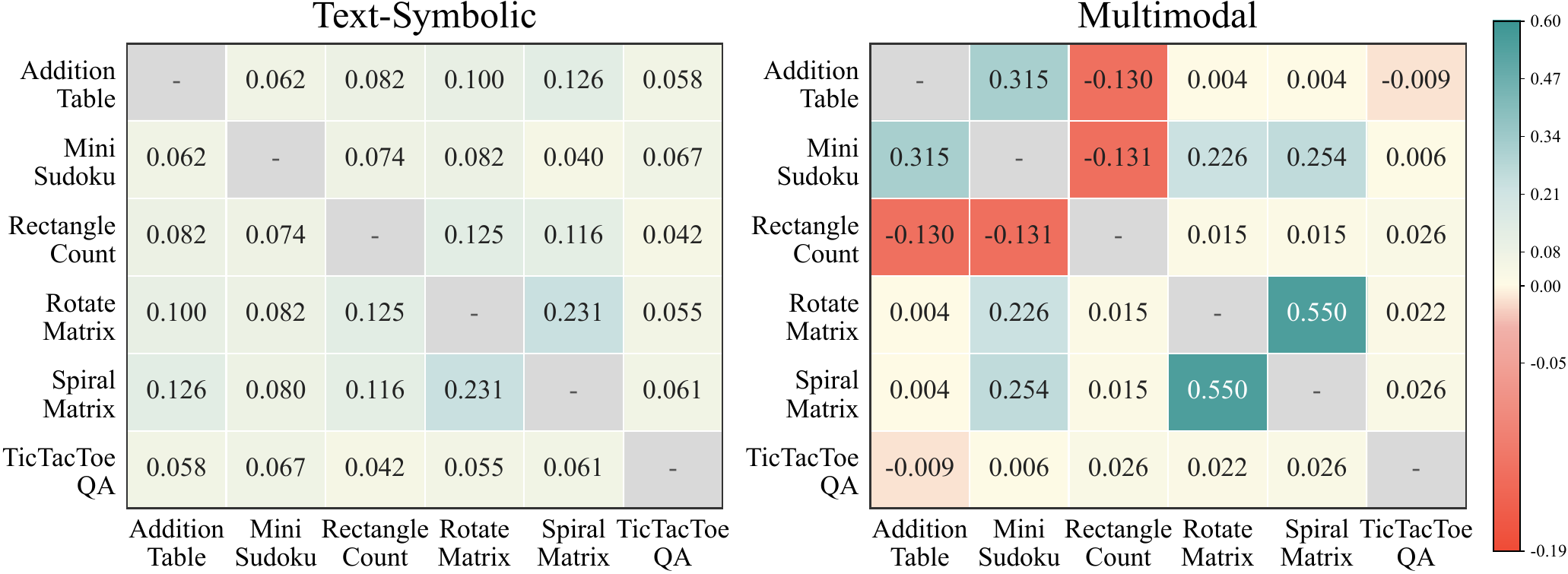}
    \caption{
    Gradient cosine similarity analysis between environments under different versions.
    }
    \label{fig:gradient_analysis}
    \vspace{-10pt}
\end{figure*}


To determine whether the mixed-training degradation simply comes from logic conflicts among environments, we conduct a controlled comparison. Specifically, we construct two versions of the same environment set: text-symbolic and multimodal. In the text-symbolic version, visual observations are converted into text observations, while other information remains unchanged. For each version, we train separate models on individual environments and also train one model on mixed-environment.

We first compare how mixed-environment training affects the two versions. As shown in Table~\ref{tab:negative_transfer_summary}, the text-symbolic version is more stable under mixed training. Compared with single-environment training, mixed training causes only a $1.3\%$ drop in the text-symbolic setting, but a much larger $10.7\%$ drop in the multimodal setting, indicating stronger negative transfer. Appendix~\ref{appendix:performance_transfer_across_modalities} provides the full transfer matrices, including cross-environment evaluation results of single-environment models. These results further show that multimodal environments are more prone to negative transfer. 

We further analyze training conflicts from the gradient perspective. Given two environments, we compute their training gradients $g_i$ and $g_j$, and use cosine similarity to measure the consistency of their update directions. A large $|\cos(g_i, g_j)|$ indicates strong coupling between the update directions, and negative values suggest conflict.
As shown in Figure~\ref{fig:gradient_analysis}. Compared with the text-symbolic version, the multimodal version exhibits more polarized gradient similarities, including several strongly negative correlations. This indicates that even when the underlying task logic remains unchanged, the multimodal environments can amplify optimization conflicts between environments. Overall, these results suggest that mixed-environment training degradation is more pronounced in multimodal environments. This motivates a closer study of how to construct effective training environment distributions for multimodal agents.

\subsection{Multimodal-Specific Failure Modes}
\label{sec:bottleneck}

\begin{figure}[t]
\centering
  \includegraphics[width=0.95\columnwidth]{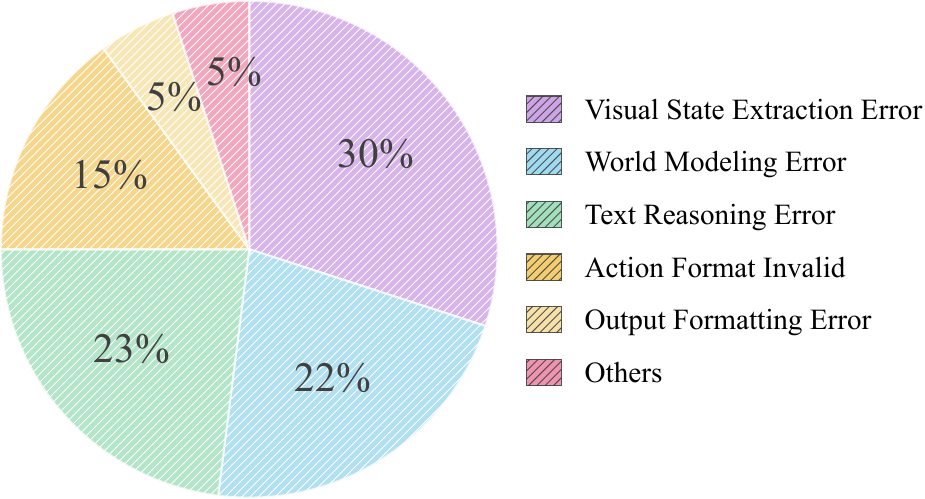}
  \caption{Error analysis of multimodal agents.}
  \label{fig:error_analysis}
    \vspace{-15pt}
\end{figure}

Finally, we analyze multimodal-specific bottlenecks of models in multimodal environments. We examine 200 failure trajectories of Qwen3-VL-4B and conduct manual error analysis. As shown in Figure ~\ref{fig:error_analysis}, we find that model failures mainly stem from two factors. The first is \emph{visual state extraction}. Models often fail to reliably extract the environment state from visual observations. The second is \emph{world modeling}. The model may fail to correctly understand environment rules and how actions affect the environment. These findings are consistent with recent studies~\cite{visgym}. However, these two bottlenecks cannot be solved simply by increasing the number of environments. Instead, it requires a difficult schedule tailored to the unique challenges of multimodal agents to help the model overcome these limitations~\cite{gym-v}. 



\subsection{From Simply Environment Scaling to Effective Environment Distribution}
\label{sec:sample_to_distribution}

The preliminary experiments show that the key to multimodal environments is not merely to construct \textbf{more ``usable'' environments}, but to construct a more \textbf{effective training environment distribution}. Therefore, we focus on how to design effective environment distributions for multimodal agents. We argue that the effectiveness of an environment distribution should be evaluated along at least two core dimensions: \textcolor{myred}{\textbf{diversity}} and \textcolor{myblue}{\textbf{difficulty structure}}. In the following sections, we will study these two dimensions and introduce corresponding methods.



\section{\textcolor{myred}{Diversity}: Environmental Selection}
\label{sec:diversity}

The first key dimension is \textcolor{myred}{\textbf{diversity}}. Existing works often measure environment diversity using surface representations such as task descriptions or environment code~\cite{EnvScaler, Agent_World_Model}. However, these representations do not necessarily reflect the actual abilities that agents need to learn in the environments. Therefore, we characterize environment diversity from the perspective of \emph{agent ability diversity}. The core idea is that the underlying diversity of environments should be analyzed through agent behaviors and the abilities required to solve the tasks. Based on this idea, we first decompose agent trajectories into a set of reusable \emph{atomic abilities}, and then build a \emph{meta-ability profile} for each environment. We further propose \textbf{Ability-aware Environment Selection (AES)}, which selects a compact and complementary environment subset based on meta-ability profiles and gradient-based conflict analysis. We describe this method in detail below. We show the complete pipeline on the left side of Figure \ref{fig:methodology_intro}.


\subsection{Construct Meta-Ability Profiles}
\label{sec:atomic_behavior}

As shown in the upper-left panel of Figure~\ref{fig:methodology_intro}, for each environment, we collect 40 agent trajectories: half from Qwen3-VL-4B and half from a stronger model, Gemini-3-Flash~\cite{gemini3.1}. Using models with different capability levels allows the trajectory pool to contain both successful strategies and failure patterns, which reveal the underlying ability requirements from complementary perspectives. We then use a strong annotation model, GPT-5~\cite{gpt-5.4}, to perform \textbf{atomic ability segmentation}. Given the environment information and model responses, GPT-5 decomposes each complete trajectory into a sequence of atomic abilities. Here, \textit{atomic} does not refer to the lowest-level tokens, but to interpretable behavior units that are meaningful for solving the task, such as ``identifying the target position''. To control the segmentation granularity, we design a unified prompt with few-shot examples and manually inspect the outputs to filter low-quality segmentations, such as overly coarse labels like ``perception''. Detailed prompts are provided in Appendix~\ref{appendix:atomic_ability_prompt}.

After obtaining atomic abilities from trajectories, we aggregate them to build a meta-ability profile $P_e$ for each environment $e$. We first use GPT-5 to merge semantically equivalent atomic abilities. Then, we compute the frequency of each meta-ability and the transition frequency between different meta-abilities, forming an ability graph. Low-frequency and unstable abilities are filtered out. Finally, each environment is represented by a \textbf{meta-ability profile}, which contains meta-abilities, transition edges, and their frequencies. We further divide meta-abilities into \textit{core and soft abilities}. Core abilities appear stably in an environment and are directly related to task completion, while soft abilities appear occasionally or are only weakly related to solving the task. Through this process, we construct meta-ability profiles for all 200 environments and summarize 72 core meta-abilities.

\subsection{Methodology: \textcolor{myred}{Ability-aware Environment Selection (AES)}}
\label{sec:bcrc}

As shown in the lower-left panel of Figure~\ref{fig:methodology_intro}, after obtaining the meta-ability profile of each environment, we propose \textbf{Ability-aware Environment Selection (AES)}. We point out that the goal of diversity should be \textbf{broader core capability coverage, with less redundancy and conflict}. 

For each candidate environment $e$, we define its coverage set $C(e)$, which is constructed from the core and soft abilities in its profile, where core abilities are assigned higher weights. Given the current selected environment set $S$, the new coverage brought by $e$ is defined as:
\[
\mathrm{NewCoverage}(e, S)=\sum_{a \in C(e)\setminus C(S)} w_a ,
\]
where $w_a$ denotes the weight of ability $a$, and $C(S)$ denotes the union of abilities already covered by the selected set $S$.

We also consider two factors that may influence joint training: \textbf{redundancy} and \textbf{conflict}. Redundancy is measured by the similarity between meta-ability profiles. If an environment is highly similar to the selected environments, it may provide limited new training information. We define:
\[
\mathrm{Redundancy}(e,S)=\max_{e' \in S} \mathrm{sim}(P_e, P_{e'}),
\]
where $\mathrm{sim}(\cdot,\cdot)$ is the weighted profile similarity.

Conflict is estimated by the gradient between environments. As discussed in Section~\ref{sec:negative_transfer}, when two environments have negative gradient cosine similarity, they are more likely to introduce optimization interference during joint training. Therefore, we define the conflict score as:
\[
\mathrm{Conflict}(e,S)=\max_{e' \in S} \max(0, -\cos(g_e,g_{e'})).
\]

Finally, in each selection step, the gain of a candidate environment $e$ is defined as:
\[
\begin{aligned}
\mathrm{Gain}(e \mid S) =
&\ \lambda_1 \mathrm{NewCoverage}(e,S) \\
&- \lambda_2 \mathrm{Redundancy}(e,S) \\
&- \lambda_3 \mathrm{Conflict}(e,S).
\end{aligned}
\]
Here, $\lambda_1,2,3$ are hyperparameters that control the relative weights.
The algorithm starts from an empty set. At each step, it selects the environment with the highest gain and adds it to $S$. This process is repeated until all core behaviors are covered.

\begin{figure}[t]
  \includegraphics[width=\columnwidth]{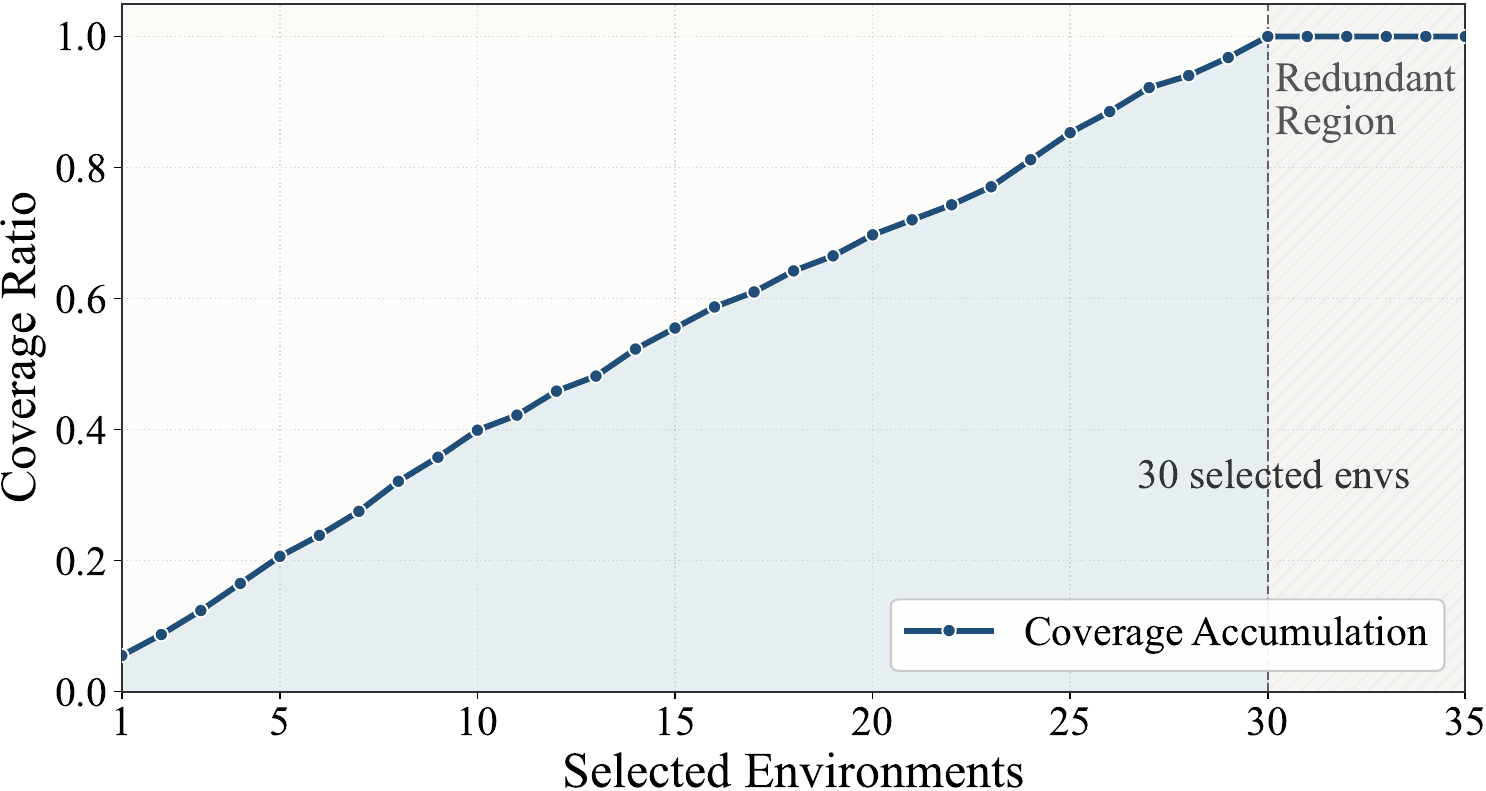}
  \caption{Core ability coverage curve when selecting environments using AES.}
  \label{fig:selected_30_coverage_curve}
    \vspace{-10pt}
\end{figure}

Based on AES, we select a diverse subset of 30 environments from the original environment pool. Figure~\ref{fig:selected_30_coverage_curve} shows the coverage curve as environments are added sequentially. The coverage of core meta-abilities increases steadily and eventually reaches full coverage, suggesting that AES can preserve broad ability coverage with only a small number of environments. In Appendix~\ref{appendix:more_results_on_diversity}, we further visualize the selected environments in profile space. 

\section{\textcolor{myblue}{Difficulty}: Curriculum Learning}
\label{sec:curriculum}

The experiments in Section \ref{sec:bottleneck} show that the bottleneck of multimodal agents comes more from the insufficient ability to extract visual states and model environmental rules. Thus when only raw visual inputs are provided at the early training stage, the model often struggles to solve the tasks, leading to sparse RL reward signals and unstable training.

Based on this, we design several textual harnesses to help models overcome the bottlenecks. In this work, a \emph{harness} refers to a type of \emph{training scaffold}: auxiliary information provided by the environment to help the model extract states from visual observations and understand rules. We consider four types of harnesses: textual observations, text states, text hints, and rule descriptions. They provide, respectively, text-symbolic descriptions of visual observations, key environment states, hints about important visual content, and explicit task rules. Examples of different harness types are shown in Figure~\ref{fig:methodology_intro}. We evaluate the effect of these harnesses in Table~\ref{tab:harness_useful}. With harness assistance, the model can solve more environments, which can provide more reliable training signals for RL.

We further use harnesses to design a two-dimensional difficulty schedule tailored to the bottlenecks of multimodal agents. Specifically, we propose \textbf{Hierarchical Difficulty Curriculum (HDC)}, which organizes training along two difficulty axes: \textbf{harness weakening} and \textbf{state-scale progression}. Harness weakening gradually removes auxiliary scaffolds, forcing the model to rely more on raw visual observations. \emph{State-scale difficulty} refers to the scale complexity of an environment instance and is controlled by environment-specific state parameters such as grid size. Our curriculum follows a hierarchical structure. State-scale progression serves as the inner curriculum, gradually increasing instance complexity within each harness stage and Harness weakening serves as the outer curriculum. Next, we will introduce them separately. 




\begin{table}[t]
\centering
\small
\setlength{\tabcolsep}{3.2pt}
\renewcommand{\arraystretch}{0.95}
\begin{tabular}{@{}lcccccc@{}}
\toprule
\multirow{2}{*}{\textbf{Harness}} 
& \multicolumn{2}{c}{\textbf{Selected Envs.}} 
& \multicolumn{2}{c}{\textbf{Held-Out Envs.}} 
& \multicolumn{2}{c}{\textbf{Avg.}} \\
\cmidrule(lr){2-3} \cmidrule(lr){4-5} \cmidrule(l){6-7}
& \textbf{ST} & \textbf{MT} & \textbf{ST} & \textbf{MT} & \textbf{ST} & \textbf{MT} \\
\midrule
None        & 13.1 & 11.9 & 13.8 & 9.5  & 13.5 & 10.7 \\
Text Obs.   & 28.2 & 18.7 & 30.6 & 17.1 & 29.4 & 17.9 \\
Text State  & \textbf{30.5} & \textbf{20.3} & \textbf{32.9} & \textbf{21.4} & \textbf{31.7} & \textbf{20.9} \\
Text Hint   & 20.7 & 15.1 & 24.6 & 13.6 & 22.7 & 14.4 \\
Rule        & 18.8 & 16.5 & 20.8 & 15.7 & 19.8 & 16.1 \\
\bottomrule
\end{tabular}
\caption{
The effects of different textual harness information.
ST and MT denote single-turn and multi-turn.
}
\label{tab:harness_useful}
\vspace{-10pt}
\end{table}

\subsection{Harness Weakening as Outer Curriculum}
\label{sec:harness_curriculum}

We first define outer curriculum axis based on harness and divide it into five difficulty levels, denoted as $H_0$ to $H_4$.
$H_0$ corresponds to full harnesses. As the level increases, harness is gradually removed. Finally, $H_4$ retains only the basic information, including raw visual observations and task descriptions. Table~\ref{tab:harness_levels} in Appendix summarizes the harnesses available at each level.

In practice, for each environment $e$, we maintain a current harness frontier: 
$r_e \in \{0,1,2,3,4\}.$ During training, samples are not always generated from $r_e$. Instead, earlier harness levels can also be sampled, which avoids abrupt distribution shifts and mitigates forgetting. For each training instance, we sample its harness level from an environment-specific distribution $D_e$:
\[
h \sim D_e(h \mid r_e).
\]
This distribution assigns a probability $p_{\mathrm{cur}}$ to the current frontier harness $r_e$.
The remaining probability mass $1-p_{\mathrm{cur}}$ is distributed over earlier harness levels $h<r_e$ according to an exponential decay:
\[
w_h = \exp\big(-\alpha (r_e-h)\big),
\]
\[
D_e(h \mid r_e) =
(1-p_{\mathrm{cur}})
\frac{w_h}{\sum_{j<r_e} w_j},
\quad h<r_e.
\]

\begin{table*}[t]
\centering
\small
\setlength{\tabcolsep}{3.8pt}
\renewcommand{\arraystretch}{1.08}

\newcommand{\acell}[1]{\cellcolor{myred!12}#1}
\newcommand{\hcell}[1]{\cellcolor{myblue!12}#1}
\newcommand{\fcell}[1]{\cellcolor{gray!22}#1}

\resizebox{\textwidth}{!}{
\begin{tabular}{lllcccccccccc}
\toprule
\multirow{2}{*}{\textbf{Model}}
& \multirow{2}{*}{\textbf{Method}}
& \multirow{2}{*}{\textbf{ID-Split}}
& \multicolumn{3}{c}{\textbf{ID Envs.}}
& \multicolumn{3}{c}{\textbf{OOD Envs.}}
& \multicolumn{4}{c}{\textbf{General Benchmarks}} \\
\cmidrule(lr){4-6}
\cmidrule(lr){7-9}
\cmidrule(lr){10-13}
& &
& \textbf{ST} & \textbf{MT} & \textbf{Rel.}
& \textbf{ST} & \textbf{MT} & \textbf{Rel.}
& \textbf{MathVision} & \textbf{MMMU} & \textbf{MMStar} & \textbf{Rel.} \\
\midrule

\multirow{7}{*}{\textbf{4B}}
& Base Model              & Random
& 15.7 & 13.1 & 0.0
& 13.8 & 9.5 & 0.0
& 52.3 & 66.0 & 62.7 & 0.0 \\

& Base Model              & \textcolor{myred}{AES}
& 13.1 & 11.9 & 0.0
& 13.8 & 9.5 & 0.0
& 52.3 & 66.0 & 62.7 & 0.0 \\

& All Envs.               & Full
& 25.4 & 19.8 & 80.1
& 17.6 & 8.3 & 7.5
& 51.6 & 66.1 & \textbf{63.6} & 0.1 \\

& Random-$K$              & Random
& 32.2 & 21.3 & 83.8
& 17.9 & 7.1 & 2.2
& 52.6 & 66.7 & 62.0 & 0.2 \\

& \hcell{Random-$K$ + \textcolor{myblue}{HDC}} & \hcell{Random}
& \hcell{37.3} & \hcell{29.4} & \hcell{131.0}
& \hcell{18.7} & \hcell{9.0} & \hcell{15.1}
& \hcell{52.0} & \hcell{65.6} & \hcell{63.9} & \hcell{0.2} \\

& \acell{\textcolor{myred}{AES}} & \acell{\textcolor{myred}{AES}}
& \acell{37.7} & \acell{25.4} & \acell{150.6}
& \acell{21.0} & \acell{12.2} & \acell{40.3}
& \acell{53.0} & \acell{67.9} & \acell{62.3} & \acell{1.2} \\

& \fcell{\textbf{\textcolor{myred}{AES} + \textcolor{myblue}{HDC}}} 
& \fcell{\textcolor{myred}{AES}}
& \fcell{\textbf{45.0}} & \fcell{\textbf{36.2}} & \fcell{\textbf{223.9}}
& \fcell{\textbf{22.1}} & \fcell{\textbf{16.8}} & \fcell{\textbf{68.5}}
& \fcell{\textbf{53.3}} & \fcell{\textbf{67.2}} & \fcell{63.1} & \fcell{\textbf{1.5}} \\

\midrule

\multirow{7}{*}{\textbf{8B}}
& Base Model              & Random
& 17.6 & 14.8 & 0.0
& 16.3 & 11.2 & 0.0
& 54.6 & 69.1 & 64.0 & 0.0 \\

& Base Model              & \textcolor{myred}{AES}
& 16.0 & 13.3 & 0.0
& 16.3 & 11.2 & 0.0
& 54.6 & 69.1 & 64.0 & 0.0 \\

& All Envs.               & Full
& 29.7 & 20.5 & 69.9
& 19.6 & 12.5 & 15.9
& \textbf{55.3} & 68.6 & \textbf{65.3} & \textbf{0.9} \\

& Random-$K$              & Random
& 34.7 & 25.9 & 86.1
& 20.8 & 9.5 & 6.2
& 54.3 & 69.9 & 64.4 & 0.4 \\

& \hcell{Random-$K$ + \textcolor{myblue}{HDC}} & \hcell{Random}
& \hcell{43.0} & \hcell{33.2} & \hcell{143.4}
& \hcell{22.3} & \hcell{10.3} & \hcell{14.4}
& \hcell{55.6} & \hcell{68.8} & \hcell{64.7} & \hcell{0.8} \\

& \acell{\textcolor{myred}{AES}} & \acell{\textcolor{myred}{AES}}
& \acell{41.0} & \acell{30.9} & \acell{144.3}
& \acell{23.2} & \acell{17.0} & \acell{47.1}
& \acell{54.0} & \acell{69.2} & \acell{65.0} & \acell{0.2} \\

& \fcell{\textbf{\textcolor{myred}{AES} + \textcolor{myblue}{HDC}}} 
& \fcell{\textcolor{myred}{AES}}
& \fcell{\textbf{49.9}} & \fcell{\textbf{40.1}} & \fcell{\textbf{206.7}}
& \fcell{\textbf{26.5}} & \fcell{\textbf{20.7}} & \fcell{\textbf{73.7}}
& \fcell{54.9} & \fcell{\textbf{70.1}} & \fcell{63.5} & \fcell{0.4} \\

\bottomrule
\end{tabular}
}
\caption{
Main results with $K=30$. ST denotes single-turn success rate, and MT denotes normalized return in multi-turn environments.
ID-Split specifies the evaluation split for the Random and \textcolor{myred}{AES} subsets.
Rel. is the relative gain over the corresponding base model.
\colorbox{gray!25}{Gray rows} denote \textbf{our complete method} and show the best performance.
}
\label{tab:main_results}

\end{table*}

\subsection{State-Scale Difficulty as Inner Curriculum}
\label{sec:scale_curriculum}

Even under a strong harness, the agent may fail to obtain rewards if it is exposed too early to complex state (e.g. complex grids). Therefore we introduce a second curriculum axis, \emph{state-scale difficulty}.

For each environment $e$, we define a set of state-scale difficulty levels: $s \in \{0,1,\ldots,S_e\}.$ The meaning of $s$ is environment-specific (e.g. grid size). And a state-scale level corresponds to a range of environment parameters. Formally, for each environment $e$ and scale level $s$, we define an environment-specific parameter distribution:
\[
\theta \sim Q_e(\theta \mid s),
\]
where $\theta$ denotes the parameters used to generate an environment instance.

During training, each environment independently maintains a current state-scale frontier $u_e$. Similar to harness sampling, we avoid training only on a single difficulty level. Instead, we sample $s$ from a sliding window of size $\Delta d$:
\[
s \sim \mathrm{Uniform}\big(\{\ell_e,\ell_e+1,\ldots,u_e\}\big),
\]
where the lower bound of the sampling window is defined as $\ell_e = \max(0, u_e - \Delta d).$




\subsection{Methodology: \textcolor{myblue}{Hierarchical Difficulty Curriculum (HDC)}}
\label{sec:nested_curriculum}
We now describe the overall algorithm. For each environment $e$, we independently maintain its curriculum state as a tuple $(r_e, \ell_e, u_e)$, where $r_e$ denotes the current harness frontier and $[\ell_e,u_e]$ denotes the current state-scale sampling window.

Each training sample is generated as follows. We first sample an environment, $e \sim \mathcal{E}_{\mathrm{train}}$. Given the current harness frontier $r_e$, we sample a harness level from $h \sim D_e(h \mid r_e)$, as described in Section~\ref{sec:harness_curriculum}. We then sample a state-scale level from the current window, $s \sim \mathrm{Uniform}(\{\ell_e,\ell_e+1,\ldots,u_e\})$. Finally, we generate a concrete training instance according to the tuple $(e,h,s)$ and optimize the policy. Importantly, environments do not need to share the same difficulty level; each environment maintains its own curriculum progress.

As shown in Figure~\ref{fig:methodology_intro}, the curriculum update is also hierarchical. When the model reaches the scale advancement threshold $\tau_s$, we advance the inner curriculum, $u_e \leftarrow u_e + 1$. When $u_e$ reaches the target state-scale level $\mathrm{target}_e$ and the model satisfies the harness advancement threshold $\tau_h$, we advance the outer curriculum, $r_e \leftarrow r_e + 1$, and then reset the state-scale. Therefore, \textbf{HDC} can be summarized as follows: within each harness level, we increase state-scale; after the model achieves sufficient competence, we weaken the harness and restart scale. Appendix~\ref{appendix:Pseudocode_of_Curriculum_Learning} and ~\ref{appendix:training_curve_of_HDC} provides the pseudocode and training curve of HDC.

\section{Experiments}
\label{sec:experiments}

We conducted extensive experiments to evaluate the effectiveness of our methods:
\textcolor{myred}{AES} and \textcolor{myblue}{HDC}.

\subsection{Experimental Setup}
\label{sec:exp_setup}

We conduct the main experiments on Qwen3-VL-4B/8B-Instruct. As a control baseline, we construct \textbf{Random-K} by sampling 30 environments from the original 200-environment pool, with environments evenly sampled across human-annotated categories inherited from the original works~\cite{visgym,gym-v}. 
We further evaluate the trained models on general multimodal benchmarks from different categories. Detailed experimental settings are provided in Appendix~\ref{appendix:exp_details}.

\subsection{Main Results}
\label{sec:main_results}

\textbf{Effect of \textcolor{myred}{Diversity-Aware Environment Selection (AES)}.}
Training on the AES-selected environments achieves stronger environment performance than both Random-$K$ and All Envs. Averaged over the ID/OOD environment groups and two model scales, AES obtains a 95.6\% relative gain over the base model, compared with 43.4\% for All Envs. This suggests that using more environments is not always better, since large environment pools may contain redundant or conflicting environments that reduce the efficiency of joint training. AES also outperforms Random-$K$, which only achieves a 44.6\% relative gain. This indicates that human-defined categories is less effective than AES which selects environments from the model's ability perspective. Meanwhile, AES shows good generalization and robustness. Although the selected environments are constructed using trajectory and gradient information from Qwen3-VL-4B, they remain effective for Qwen3-VL-8B, where AES achieves a 144.3\% relative gain on ID environments and a 47.1\% relative gain on OOD environments. Appendix \ref{appendix:surface_selection} further illustrates the comparison between AES and existing diversity methods.

\textbf{Effect of \textcolor{myblue}{Difficulty-Aware Curriculum Learning (HDC)}.}
HDC further improves performance across different environment sets. The strongest result is achieved by combining HDC with AES, where AES + HDC obtains an average relative gain of 143.2\% over the base model across the ID/OOD environment groups and two model scales. This shows that diversity and difficulty are not only two important dimensions of environment distribution, but can also complement each other to further improve training effectiveness. HDC also improves Random-$K$, increasing its average relative gain from 44.6\% to 73.7\%. This suggests that appropriate difficulty scheduling can partially mitigate the instability caused by heterogeneous multimodal environments. By starting from easier instances with stronger harnesses, the model can obtain more reliable learning signals, thereby alleviating imbalance among environments to some extent.

\subsection{Ablation Study}
\label{sec:ablation}

\textbf{Ablation on \textcolor{myred}{AES}.}
AES consists of three key factors: meta-ability coverage, redundancy and conflict control. To examine the effects of redundancy and conflict control, we ablate these two components separately. All variants are trained without HDC. As shown in Table~\ref{tab:ablation_selection}, the full AES achieves the largest relative gain 40.3\% over the base model on OOD environments. Removing redundancy control reduces the relative gain from 40.3\% to 25.3\%, while removing conflict control reduces it to only 2.8\%. These results show that both components are important for our AES method.

\begin{table}[t]
\centering
\small
\setlength{\tabcolsep}{5.5pt}
\renewcommand{\arraystretch}{1.02}
\begin{tabular}{@{}lcccc@{}}
\toprule
\textbf{Method} 
& \textbf{\#Env} 
& \textbf{OOD-ST} 
& \textbf{OOD-MT} 
& \textbf{Rel.} \\
\midrule
Base Model          
& -- & 13.8 & 9.5  & 0.0  \\

Random-$K$          
& 30 & 17.9 & 7.1  & 2.2  \\

\textcolor{myred}{AES} w/o Redund.     
& 41 & 18.6 & 11.0 & 25.3 \\

\textcolor{myred}{AES} w/o Conflict    
& 34 & 18.5 & 6.8  & 2.8  \\

\rowcolor{gray!12}
\textbf{\textcolor{myred}{AES}}                 
& \textbf{30} 
& \textbf{21.0} 
& \textbf{12.2} 
& \textbf{40.3} \\
\bottomrule
\end{tabular}
\caption{
Results of ablation experiments on AES.
}
\label{tab:ablation_selection}
\vspace{-10pt}
\end{table}

\textbf{Ablation on \textcolor{myblue}{HDC}.}
We ablate the two axes of HDC: harness weakening and state-scale progression. All variants use the same AES-selected environment set. As shown in Table~\ref{tab:ablation_curriculum}, both single-axis curricula benefit. The scale-only curriculum brings an average relative gain of 11.5\%, while the harness-only curriculum brings a larger gain of 18.1\%. Combining both axes achieves the best performance, with an average relative gain of 27.7\%. These results show that harness weakening and state-scale progression are complementary and jointly improve multimodal environment training.

\begin{table}[t]
\centering
\small
\setlength{\tabcolsep}{4.0pt}
\renewcommand{\arraystretch}{1.02}
\begin{tabular}{@{}lccccc@{}}
\toprule
\multirow{2}{*}{\textbf{Method}} 
& \multicolumn{2}{c}{\textbf{ID Ens.}} 
& \multicolumn{2}{c}{\textbf{OOD Ens.}} 
& \multirow{2}{*}{\textbf{Rel.}} \\
\cmidrule(lr){2-3} \cmidrule(lr){4-5}
& \textbf{ST} & \textbf{MT} 
& \textbf{ST} & \textbf{MT} 
& \\
\midrule

Base Model         
& 13.1 & 11.9 & 13.8 & 9.5  & -- \\

\textcolor{myred}{AES}                
& 37.7 & 25.4 & 21.0 & 12.2 & 0.0 \\

\textcolor{myred}{AES} + Scale-only   
& 40.7 & 29.8 & 21.6 & 14.1 & 11.5 \\

\textcolor{myred}{AES} + Harness-only 
& 41.9 & 33.1 & 21.2 & 15.3 & 18.1 \\

\rowcolor{gray!12}
\textbf{\textcolor{myred}{AES} + \textcolor{myblue}{HDC}}          
& \textbf{45.0} & \textbf{36.2} 
& \textbf{22.1} & \textbf{16.8} 
& \textbf{27.7} \\

\bottomrule
\end{tabular}
\caption{
Results of ablation experiments on HDC.
}
\label{tab:ablation_curriculum}
\vspace{-10pt}
\end{table}

\section{Related Works}
\label{sec:related_works}

We summarize the research most relevant to our core ideas, and provide more discussions in Appendix~\ref{appendix:more_related_works}.
Recently, multimodal environmental and agent learning have become cutting-edge research areas~\cite{odysseus}. Among them, we mainly focused on general multimodal environments~\cite{gym-v, visgym}.
Another important line of work studies environment quality verification~\cite{R2E-Gym}. They focused on whether environments are executable~\cite{SWE-Gym}, whether state transitions are correct and so on~\cite{Agent2World, AutoWebWorld}. These approaches mainly address sample-level quality. However, how to assess the effectiveness of environment distribution remains less systematically studied~\cite{huang2025scaling}. In particular, for multimodal environments, the diversity and difficulty structure  remain insufficiently explored. Current diversity control methods often rely on task-description embeddings~\cite{V-GameGym} or tool categories~\cite{Api-bank}. Meanwhile, difficulty control in multimodal environments largely follows the framework of text-based environments~\cite{RLVE,gg-bench, AI_GAMESTORE}. To bridge this gap, we analyze diversity and difficulty separately and propose corresponding methods.

\section{Conclusion}

In this paper, we show that simply scaling multimodal environments is not always effective for agent training. Our analyses reveal that the effectiveness of an environment distribution depends on both diversity and difficulty structure. Based on this insight, we propose Ability-aware Environment Selection (AES) to construct diverse environment set, and Hierarchical Difficulty Curriculum (HDC) specific to multimodal agents. Experiments demonstrate that AES and HDC consistently improve agent training.

\section*{Limitations}

Although we have conducted extensive experiments and analyses, there are still some aspect that need improvement. First, due to cost constraints, our environment pool is mainly built upon existing multimodal environment works, and we do not further study large-scale environment synthesis. This remains an important direction for future exploration. Second, due to limited computational resources, we mainly compare different methods under a unified compute budget. Since multimodal environment training, especially multi-turn training, is computationally expensive, some environments may not be fully trained. Future work can further validate our findings under larger training budgets. Finally, AES uses gradient information to characterize optimization conflicts between environments, which introduces additional offline computation compared with traditional diversity measures. Future work may explore more efficient conflict estimation methods or alternative metrics.


\section*{Ethics and Artifact Use Statement}

\textbf{Potential risks.}
This work studies how to design effective training environment distributions for multimodal agent learning. A potential risk is that environment-based training may improve agents' ability of decision making in interactive tasks. If such models are deployed in open real-world systems without sufficient safeguards, they may produce incorrect actions or other unintended effects. Therefore, the models trained in this work are evaluated only in controlled and verifiable multimodal research environments, and should not be directly deployed in open real-world systems without proper safety treatment.

\textbf{Artifacts, licenses, and intended use.}
Our environment pool is mainly built upon existing multimodal environment works and is used only for research purposes. We properly cite all assets and adhere to their licenses and terms of use.

\textbf{Data privacy and content safety.}
The environments used in this work are virtual multimodal interactive environments and do not involve collecting personal data from real users.

\textbf{Use of AI assistants.}
First, in our method, LLMs were used for atomic ability segmentation and environment profile construction  with manual inspection by the authors. Second, during writing, LLMs were used to check grammar and polish text, and we carefully verified that their use did not alter the original meaning. In addition, AI assistants were used to improve the presentation of some tables and figures, while all experimental results were checked by the authors and remained unchanged.



\bibliography{custom}

@article{visgym,
  author       = {Zirui Wang and
                  Junyi Zhang and
                  Jiaxin Ge and
                  Long Lian and
                  Letian Fu and
                  Lisa Dunlap and
                  Ken Goldberg and
                  Xudong Wang and
                  Ion Stoica and
                  David M. Chan and
                  Sewon Min and
                  Joseph E. Gonzalez},
  title        = {VisGym: Diverse, Customizable, Scalable Environments for Multimodal
                  Agents},
  journal      = {CoRR},
  volume       = {abs/2601.16973},
  year         = {2026},
  url          = {https://doi.org/10.48550/arXiv.2601.16973},
  doi          = {10.48550/ARXIV.2601.16973},
  eprinttype   = {arXiv},
  eprint       = {2601.16973},
  bibsource    = {dblp computer science bibliography, https://dblp.org}
}

@article{gym-v,
  author       = {Fanqing Meng and
                  Lingxiao Du and
                  Jiawei Gu and
                  Jiaqi Liao and
                  Linjie Li and
                  Zijian Wu and
                  Xiangyan Liu and
                  Ziqi Zhao and
                  Mengkang Hu and
                  Yue Zhang and
                  Zichen Liu and
                  Jiaheng Zhang and
                  Michael Qizhe Shieh},
  title        = {Gym-V: {A} Unified Vision Environment System for Agentic Vision Research},
  journal      = {CoRR},
  volume       = {abs/2603.15432},
  year         = {2026},
  url          = {https://doi.org/10.48550/arXiv.2603.15432},
  doi          = {10.48550/ARXIV.2603.15432},
  eprinttype   = {arXiv},
  eprint       = {2603.15432},
  bibsource    = {dblp computer science bibliography, https://dblp.org}
}

@article{odysseus,
  title={Odysseus: Scaling VLMs to 100+ Turn Decision-Making in Games via Reinforcement Learning},
  author={Shi, Chengshuai and Li, Wenzhe and Liang, Xinran and Lu, Yizhou and Yang, Wenjia and Feng, Ruirong and Karten, Seth and Yang, Ziran and Ding, Zihan and Sarch, Gabriel and others},
  journal={arXiv preprint arXiv:2605.00347},
  year={2026}
}

@article{V-GameGym,
  author       = {Wei Zhang and
                  Jack Yang and
                  Renshuai Tao and
                  Lingzheng Chai and
                  Shawn Guo and
                  Jiajun Wu and
                  Xiaoming Chen and
                  Ganqu Cui and
                  Ning Ding and
                  Xander Xu and
                  Hu Wei and
                  Bowen Zhou},
  title        = {V-GameGym: Visual Game Generation for Code Large Language Models},
  journal      = {CoRR},
  volume       = {abs/2509.20136},
  year         = {2025},
  url          = {https://doi.org/10.48550/arXiv.2509.20136},
  doi          = {10.48550/ARXIV.2509.20136},
  eprinttype   = {arXiv},
  eprint       = {2509.20136},
  bibsource    = {dblp computer science bibliography, https://dblp.org}
}

@article{AutoWebWorld,
  author       = {Yifan Wu and
                  Yiran Peng and
                  Yiyu Chen and
                  Jianhao Ruan and
                  Zijie Zhuang and
                  Cheng Yang and
                  Jiayi Zhang and
                  Man Chen and
                  Yenchi Tseng and
                  Zhaoyang Yu and
                  Liang Chen and
                  Yuyao Zhai and
                  Bang Liu and
                  Chenglin Wu and
                  Yuyu Luo},
  title        = {AutoWebWorld: Synthesizing Infinite Verifiable Web Environments via
                  Finite State Machines},
  journal      = {CoRR},
  volume       = {abs/2602.14296},
  year         = {2026},
  url          = {https://doi.org/10.48550/arXiv.2602.14296},
  doi          = {10.48550/ARXIV.2602.14296},
  eprinttype   = {arXiv},
  eprint       = {2602.14296},
  bibsource    = {dblp computer science bibliography, https://dblp.org}
}

@article{Matrix-game,
  title={Matrix-game 2.0: An open-source real-time and streaming interactive world model},
  author={He, Xianglong and Peng, Chunli and Liu, Zexiang and Wang, Boyang and Zhang, Yifan and Cui, Qi and Kang, Fei and Jiang, Biao and An, Mengyin and Ren, Yangyang and others},
  journal={arXiv preprint arXiv:2508.13009},
  year={2025}
}

@article{Agent_World_Model,
  author       = {Zhaoyang Wang and
                  Canwen Xu and
                  Boyi Liu and
                  Yite Wang and
                  Siwei Han and
                  Zhewei Yao and
                  Huaxiu Yao and
                  Yuxiong He},
  title        = {Agent World Model: Infinity Synthetic Environments for Agentic Reinforcement
                  Learning},
  journal      = {CoRR},
  volume       = {abs/2602.10090},
  year         = {2026},
  url          = {https://doi.org/10.48550/arXiv.2602.10090},
  doi          = {10.48550/ARXIV.2602.10090},
  eprinttype   = {arXiv},
  eprint       = {2602.10090},
  bibsource    = {dblp computer science bibliography, https://dblp.org}
}

@article{EnvScaler,
  author       = {Xiaoshuai Song and
                  Haofei Chang and
                  Guanting Dong and
                  Yutao Zhu and
                  Zhicheng Dou and
                  Ji{-}Rong Wen},
  title        = {EnvScaler: Scaling Tool-Interactive Environments for {LLM} Agent via
                  Programmatic Synthesis},
  journal      = {CoRR},
  volume       = {abs/2601.05808},
  year         = {2026},
  url          = {https://doi.org/10.48550/arXiv.2601.05808},
  doi          = {10.48550/ARXIV.2601.05808},
  eprinttype    = {arXiv},
  eprint       = {2601.05808},
  bibsource    = {dblp computer science bibliography, https://dblp.org}
}

@online{gemini3.1,
  author = {{Google DeepMind}},
  title = {Gemini 3.1 Pro},
  year = {2025},
  url = {https://deepmind.google/models/gemini/pro/},
  note = {Accessed: 2026-03-16}
}

@online{gpt-5.4,
  author    = {{OpenAI}},
  title     = {Introducing GPT-5.4},
  year      = {2025},
  url       = {https://openai.com/index/introducing-gpt-5-4/},
  note      = {Accessed: 2026-03-16}
}

@article{RLVE,
  author       = {Zhiyuan Zeng and
                  Hamish Ivison and
                  Yiping Wang and
                  Lifan Yuan and
                  Shuyue Stella Li and
                  Zhuorui Ye and
                  Siting Li and
                  Jacqueline He and
                  Runlong Zhou and
                  Tong Chen and
                  Chenyang Zhao and
                  Yulia Tsvetkov and
                  Simon Shaolei Du and
                  Natasha Jaques and
                  Hao Peng and
                  Pang Wei Koh and
                  Hannaneh Hajishirzi},
  title        = {{RLVE:} Scaling Up Reinforcement Learning for Language Models with
                  Adaptive Verifiable Environments},
  journal      = {CoRR},
  volume       = {abs/2511.07317},
  year         = {2025},
  url          = {https://doi.org/10.48550/arXiv.2511.07317},
  doi          = {10.48550/ARXIV.2511.07317},
  eprinttype    = {arXiv},
  eprint       = {2511.07317},
  bibsource    = {dblp computer science bibliography, https://dblp.org}
}

@article{Qwen3,
  author       = {Qwen Team},
  title        = {Qwen3 Technical Report},
  journal      = {CoRR},
  volume       = {abs/2505.09388},
  year         = {2025},
  url          = {https://doi.org/10.48550/arXiv.2505.09388},
  doi          = {10.48550/ARXIV.2505.09388},
  eprinttype   = {arXiv},
  eprint       = {2505.09388},
  bibsource    = {dblp computer science bibliography, https://dblp.org}
}

@article{InfiniteWeb,
  author       = {Ziyun Zhang and
                  Zezhou Wang and
                  Xiaoyi Zhang and
                  Zongyu Guo and
                  Jiahao Li and
                  Bin Li and
                  Yan Lu},
  title        = {InfiniteWeb: Scalable Web Environment Synthesis for {GUI} Agent Training},
  journal      = {CoRR},
  volume       = {abs/2601.04126},
  year         = {2026},
  url          = {https://doi.org/10.48550/arXiv.2601.04126},
  doi          = {10.48550/ARXIV.2601.04126},
  eprinttype    = {arXiv},
  eprint       = {2601.04126},
  bibsource    = {dblp computer science bibliography, https://dblp.org}
}

@article{GameWorld,
  author       = {Mingyu Ouyang and
                  Siyuan Hu and
                  Kevin Qinghong Lin and
                  Hwee Tou Ng and
                  Mike Zheng Shou},
  title        = {GameWorld: Towards Standardized and Verifiable Evaluation of Multimodal
                  Game Agents},
  journal      = {CoRR},
  volume       = {abs/2604.07429},
  year         = {2026},
  url          = {https://doi.org/10.48550/arXiv.2604.07429},
  doi          = {10.48550/ARXIV.2604.07429},
  eprinttype   = {arXiv},
  eprint       = {2604.07429},
  bibsource    = {dblp computer science bibliography, https://dblp.org}
}

@article{PokeGym,
  author       = {Ruizhi Zhang and
                  Ye Huang and
                  Yuangang Pan and
                  Chuanfu Shen and
                  Zhilin Liu and
                  Ting Xie and
                  Wen Li and
                  Lixin Duan},
  title        = {PokeGym: {A} Visually-Driven Long-Horizon Benchmark for Vision-Language
                  Models},
  journal      = {CoRR},
  volume       = {abs/2604.08340},
  year         = {2026},
  url          = {https://doi.org/10.48550/arXiv.2604.08340},
  doi          = {10.48550/ARXIV.2604.08340},
  eprinttype   = {arXiv},
  eprint       = {2604.08340},
  bibsource    = {dblp computer science bibliography, https://dblp.org}
}

@article{R2E-Gym,
  author       = {Naman Jain and
                  Jaskirat Singh and
                  Manish Shetty and
                  Liang Zheng and
                  Koushik Sen and
                  Ion Stoica},
  title        = {R2E-Gym: Procedural Environments and Hybrid Verifiers for Scaling
                  Open-Weights {SWE} Agents},
  journal      = {CoRR},
  volume       = {abs/2504.07164},
  year         = {2025},
  url          = {https://doi.org/10.48550/arXiv.2504.07164},
  doi          = {10.48550/ARXIV.2504.07164},
  eprinttype    = {arXiv},
  eprint       = {2504.07164},
  bibsource    = {dblp computer science bibliography, https://dblp.org}
}

@article{SWE-smith,
  author       = {John Yang and
                  Kilian Leret and
                  Carlos E. Jimenez and
                  Alexander Wettig and
                  Kabir Khandpur and
                  Yanzhe Zhang and
                  Binyuan Hui and
                  Ofir Press and
                  Ludwig Schmidt and
                  Diyi Yang},
  title        = {SWE-smith: Scaling Data for Software Engineering Agents},
  journal      = {CoRR},
  volume       = {abs/2504.21798},
  year         = {2025},
  url          = {https://doi.org/10.48550/arXiv.2504.21798},
  doi          = {10.48550/ARXIV.2504.21798},
  eprinttype    = {arXiv},
  eprint       = {2504.21798},
  bibsource    = {dblp computer science bibliography, https://dblp.org}
}

@inproceedings{SWE-Gym,
  author       = {Jiayi Pan and
                  Xingyao Wang and
                  Graham Neubig and
                  Navdeep Jaitly and
                  Heng Ji and
                  Alane Suhr and
                  Yizhe Zhang},
  editor       = {Aarti Singh and
                  Maryam Fazel and
                  Daniel Hsu and
                  Simon Lacoste{-}Julien and
                  Felix Berkenkamp and
                  Tegan Maharaj and
                  Kiri Wagstaff and
                  Jerry Zhu},
  title        = {Training Software Engineering Agents and Verifiers with SWE-Gym},
  booktitle    = {Forty-second International Conference on Machine Learning, {ICML}
                  2025, Vancouver, BC, Canada, July 13-19, 2025},
  series       = {Proceedings of Machine Learning Research},
  volume       = {267},
  publisher    = {{PMLR} / OpenReview.net},
  year         = {2025},
  url          = {https://proceedings.mlr.press/v267/pan25g.html},
  bibsource    = {dblp computer science bibliography, https://dblp.org}
}

@inproceedings{Api-bank,
  title={Api-bank: A comprehensive benchmark for tool-augmented llms},
  author={Li, Minghao and Zhao, Yingxiu and Yu, Bowen and Song, Feifan and Li, Hangyu and Yu, Haiyang and Li, Zhoujun and Huang, Fei and Li, Yongbin},
  booktitle={Proceedings of the 2023 conference on empirical methods in natural language processing},
  pages={3102--3116},
  year={2023}
}

@inproceedings{WebArena,  author       = {Shuyan Zhou and                  Frank F. Xu and                  Hao Zhu and                  Xuhui Zhou and                  Robert Lo and                  Abishek Sridhar and                  Xianyi Cheng and                  Tianyue Ou and                  Yonatan Bisk and                  Daniel Fried and                  Uri Alon and                  Graham Neubig},  title        = {WebArena: {A} Realistic Web Environment for Building Autonomous Agents},  booktitle    = {The Twelfth International Conference on Learning Representations,                  {ICLR} 2024, Vienna, Austria, May 7-11, 2024},  publisher    = {OpenReview.net},  year         = {2024},  url          = {https://openreview.net/forum?id=oKn9c6ytLx},  bibsource    = {dblp computer science bibliography, https://dblp.org}}

@article{LOGIGEN,
  title={LOGIGEN: Logic-Driven Generation of Verifiable Agentic Tasks},
  author={Zeng, Yucheng and Lu, Weipeng and Liu, Linyun and Li, Shupeng and Qu, Zitian and Zhu, Chenghao and Li, Shaofei and Tan, Zhengdong and Liu, Mengyue and Zhao, Haotian and others},
  journal={arXiv preprint arXiv:2603.00540},
  year={2026}
}

@article{gg-bench,
  author       = {Vivek Verma and
                  David Huang and
                  William Chen and
                  Daniel Klein and
                  Nicholas Tomlin},
  title        = {Measuring General Intelligence with Generated Games},
  journal      = {CoRR},
  volume       = {abs/2505.07215},
  year         = {2025},
  url          = {https://doi.org/10.48550/arXiv.2505.07215},
  doi          = {10.48550/ARXIV.2505.07215},
  eprinttype    = {arXiv},
  eprint       = {2505.07215},
  bibsource    = {dblp computer science bibliography, https://dblp.org}
}

@misc{AI_GAMESTORE,
      title={AI Gamestore: Scalable, Open-Ended Evaluation of Machine General Intelligence with Human Games}, 
      author={Lance Ying and Ryan Truong and Prafull Sharma and Kaiya Ivy Zhao and Nathan Cloos and Kelsey R. Allen and Thomas L. Griffiths and Katherine M. Collins and José Hernández-Orallo and Phillip Isola and Samuel J. Gershman and Joshua B. Tenenbaum},
      year={2026},
      eprint={2602.17594},
      archivePrefix={arXiv},
      primaryClass={cs.AI},
      url={https://arxiv.org/abs/2602.17594}, 
}

@article{e2h_reasoner,
  title={Curriculum reinforcement learning from easy to hard tasks improves LLM reasoning},
  author={Parashar, Shubham and Gui, Shurui and Li, Xiner and Ling, Hongyi and Vemuri, Sushil and Olson, Blake and Li, Eric and Zhang, Yu and Caverlee, James and Kalathil, Dileep and others},
  journal={arXiv preprint arXiv:2506.06632},
  year={2025}
}

@inproceedings{ADCL,
  title={Learning like humans: Advancing llm reasoning capabilities via adaptive difficulty curriculum learning and expert-guided self-reformulation},
  author={Zhang, Enci and Yan, Xingang and Lin, Wei and Zhang, Tianxiang and Qianchun, Lu},
  booktitle={Proceedings of the 2025 Conference on Empirical Methods in Natural Language Processing},
  pages={6630--6644},
  year={2025}
}

@article{Vcrl,
  title={Vcrl: Variance-based curriculum reinforcement learning for large language models},
  author={Jiang, Guochao and Feng, Wenfeng and Quan, Guofeng and Hao, Chuzhan and Zhang, Yuewei and Liu, Guohua and Wang, Hao},
  journal={arXiv preprint arXiv:2509.19803},
  year={2025}
}

@inproceedings{Adacurl,
  title={Adacurl: Adaptive curriculum reinforcement learning with invalid sample mitigation and historical revisiting},
  author={Li, Renda and Huang, Hailang and Wei, Fei and Xiong, Feng and Wang, Yong and Chu, Xiangxiang},
  booktitle={Proceedings of the AAAI Conference on Artificial Intelligence},
  volume={40},
  number={27},
  pages={23123--23131},
  year={2026}
}

@inproceedings{webrl,
  title={Webrl: Training llm web agents via self-evolving online curriculum reinforcement learning},
  author={Qi, Zehan and Liu, Xiao and Iong, Iat Long and Lai, Hanyu and Sun, Xueqiao and Sun, Jiadai and Yang, Xinyue and Yang, Yu and Yao, Shuntian and Xu, Wei and others},
  booktitle={International Conference on Learning Representations},
  volume={2025},
  pages={79791--79821},
  year={2025}
}

@article{Agent2World,
  author       = {Mengkang Hu and
                  Bowei Xia and
                  Yuran Wu and
                  Ailing Yu and
                  Yude Zou and
                  Qiguang Chen and
                  Shijian Wang and
                  Jiarui Jin and
                  Kexin Li and
                  Wenxiang Jiao and
                  Yuan Lu and
                  Ping Luo},
  title        = {Agent2World: Learning to Generate Symbolic World Models via Adaptive
                  Multi-Agent Feedback},
  journal      = {CoRR},
  volume       = {abs/2512.22336},
  year         = {2025},
  url          = {https://doi.org/10.48550/arXiv.2512.22336},
  doi          = {10.48550/ARXIV.2512.22336},
  eprinttype    = {arXiv},
  eprint       = {2512.22336},
  bibsource    = {dblp computer science bibliography, https://dblp.org}
}

@article{AutoForge,
  author       = {Shihao Cai and
                  Runnan Fang and
                  Jialong Wu and
                  Baixuan Li and
                  Xinyu Wang and
                  Yong Jiang and
                  Liangcai Su and
                  Liwen Zhang and
                  Wenbiao Yin and
                  Zhen Zhang and
                  Fuli Feng and
                  Pengjun Xie and
                  Xiaobin Wang},
  title        = {AutoForge: Automated Environment Synthesis for Agentic Reinforcement
                  Learning},
  journal      = {CoRR},
  volume       = {abs/2512.22857},
  year         = {2025},
  url          = {https://doi.org/10.48550/arXiv.2512.22857},
  doi          = {10.48550/ARXIV.2512.22857},
  eprinttype    = {arXiv},
  eprint       = {2512.22857},
  bibsource    = {dblp computer science bibliography, https://dblp.org}
}

@inproceedings{huang2025scaling,
  title={Scaling Environments for LLM Agents in the Era of Learning from Interaction: A Survey},
  author={Huang, Yuchen and Li, Sijia and Liu, Wei and Fan, Zhiyuan and Fung, Yi R and others},
  booktitle={Workshop on Scaling Environments for Agents},
  year={2025}
}

@article{Gym-Anything,
  author       = {Pranjal Aggarwal and
                  Graham Neubig and
                  Sean Welleck},
  title        = {Gym-Anything: Turn any Software into an Agent Environment},
  journal      = {CoRR},
  volume       = {abs/2604.06126},
  year         = {2026},
  url          = {https://doi.org/10.48550/arXiv.2604.06126},
  doi          = {10.48550/ARXIV.2604.06126},
  eprinttype   = {arXiv},
  eprint       = {2604.06126},
  bibsource    = {dblp computer science bibliography, https://dblp.org}
}

@article{GAME-RL,
  author       = {Dazhi Zhan and
                  Xin Liu and
                  Wei Bai and
                  Wei Li and
                  Shize Guo and
                  Zhisong Pan},
  title        = {{GAME-RL:} Generating Adversarial Malware Examples Against {API} Call
                  Based Detection via Reinforcement Learning},
  journal      = {{IEEE} Trans. Dependable Secur. Comput.},
  volume       = {22},
  number       = {5},
  pages        = {5431--5447},
  year         = {2025},
  url          = {https://doi.org/10.1109/TDSC.2025.3566708},
  doi          = {10.1109/TDSC.2025.3566708},
  bibsource    = {dblp computer science bibliography, https://dblp.org}
}

@article{VAGEN,
  author       = {Kangrui Wang and
                  Pingyue Zhang and
                  Zihan Wang and
                  Yaning Gao and
                  Linjie Li and
                  Qineng Wang and
                  Hanyang Chen and
                  Chi Wan and
                  Yiping Lu and
                  Zhengyuan Yang and
                  Lijuan Wang and
                  Ranjay Krishna and
                  Jiajun Wu and
                  Li Fei{-}Fei and
                  Yejin Choi and
                  Manling Li},
  title        = {{VAGEN:} Reinforcing World Model Reasoning for Multi-Turn {VLM} Agents},
  journal      = {CoRR},
  volume       = {abs/2510.16907},
  year         = {2025},
  url          = {https://doi.org/10.48550/arXiv.2510.16907},
  doi          = {10.48550/ARXIV.2510.16907},
  eprinttype   = {arXiv},
  eprint       = {2510.16907},
  bibsource    = {dblp computer science bibliography, https://dblp.org}
}

@article{ARC-AGI-3,
  title={ARC-AGI-3: A New Challenge for Frontier Agentic Intelligence},
  author={Foundation, ARC},
  journal={arXiv preprint arXiv:2603.24621},
  year={2026}
}

@article{SpatialEvo,
  title={SpatialEvo: Self-Evolving Spatial Intelligence via Deterministic Geometric Environments},
  author={Li, Dinging and Zhao, Yingxiu and Cheng, Xinrui and Lin, Kangheng and Peng, Hongbo and Li, Hongxing and Wang, Zixuan and Dai, Yuhong and Li, Haodong and Wang, Jia and others},
  journal={arXiv preprint arXiv:2604.14144},
  year={2026}
}

@article{Dreamgen,
  title={Dreamgen: Unlocking generalization in robot learning through video world models},
  author={Jang, Joel and Ye, Seonghyeon and Lin, Zongyu and Xiang, Jiannan and Bjorck, Johan and Fang, Yu and Hu, Fengyuan and Huang, Spencer and Kundalia, Kaushil and Lin, Yen-Chen and others},
  journal={arXiv preprint arXiv:2505.12705},
  year={2025}
}

@article{Intercode,
  title={Intercode: Standardizing and benchmarking interactive coding with execution feedback},
  author={Yang, John and Prabhakar, Akshara and Narasimhan, Karthik and Yao, Shunyu},
  journal={Advances in Neural Information Processing Systems},
  volume={36},
  pages={23826--23854},
  year={2023}
}

@article{SWE-World,
  author       = {Shuang Sun and
                  Huatong Song and
                  Lisheng Huang and
                  Jinhao Jiang and
                  Ran Le and
                  Zhihao Lv and
                  Zongchao Chen and
                  Yiwen Hu and
                  Wenyang Luo and
                  Wayne Xin Zhao and
                  Yang Song and
                  Hongteng Xu and
                  Tao Zhang and
                  Ji{-}Rong Wen},
  title        = {SWE-World: Building Software Engineering Agents in Docker-Free Environments},
  journal      = {CoRR},
  volume       = {abs/2602.03419},
  year         = {2026},
  url          = {https://doi.org/10.48550/arXiv.2602.03419},
  doi          = {10.48550/ARXIV.2602.03419},
  eprinttype   = {arXiv},
  eprint       = {2602.03419},
  bibsource    = {dblp computer science bibliography, https://dblp.org}
}

@article{Mind2web,
  title={Mind2web: Towards a generalist agent for the web},
  author={Deng, Xiang and Gu, Yu and Zheng, Boyuan and Chen, Shijie and Stevens, Sam and Wang, Boshi and Sun, Huan and Su, Yu},
  journal={Advances in Neural Information Processing Systems},
  volume={36},
  pages={28091--28114},
  year={2023}
}

@article{Mcp-universe,
  title={Mcp-universe: Benchmarking large language models with real-world model context protocol servers},
  author={Luo, Ziyang and Shen, Zhiqi and Yang, Wenzhuo and Zhao, Zirui and Jwalapuram, Prathyusha and Saha, Amrita and Sahoo, Doyen and Savarese, Silvio and Xiong, Caiming and Li, Junnan},
  journal={arXiv preprint arXiv:2508.14704},
  year={2025}
}

@article{OSWorld-MCP,
  author       = {Hongrui Jia and
                  Jitong Liao and
                  Xi Zhang and
                  Haiyang Xu and
                  Tianbao Xie and
                  Chaoya Jiang and
                  Ming Yan and
                  Si Liu and
                  Wei Ye and
                  Fei Huang},
  title        = {OSWorld-MCP: Benchmarking {MCP} Tool Invocation In Computer-Use Agents},
  journal      = {CoRR},
  volume       = {abs/2510.24563},
  year         = {2025},
  url          = {https://doi.org/10.48550/arXiv.2510.24563},
  doi          = {10.48550/ARXIV.2510.24563},
  eprinttype   = {arXiv},
  eprint       = {2510.24563},
  bibsource    = {dblp computer science bibliography, https://dblp.org}
}

@article{MobileDreamer,
  title={MobileDreamer: Generative Sketch World Model for GUI Agent},
  author={Cao, Yilin and Zhong, Yufeng and Zeng, Zhixiong and Zheng, Liming and Huang, Jing and Qiu, Haibo and Shi, Peng and Mao, Wenji and Guanglu, Wan},
  journal={arXiv preprint arXiv:2601.04035},
  year={2026}
}

@article{VLWM,
  author       = {Delong Chen and
                  Th{\'{e}}o Moutakanni and
                  Willy Chung and
                  Yejin Bang and
                  Ziwei Ji and
                  Allen Bolourchi and
                  Pascale Fung},
  title        = {Planning with Reasoning using Vision Language World Model},
  journal      = {CoRR},
  volume       = {abs/2509.02722},
  year         = {2025},
  url          = {https://doi.org/10.48550/arXiv.2509.02722},
  doi          = {10.48550/ARXIV.2509.02722},
  eprinttype   = {arXiv},
  eprint       = {2509.02722},
  bibsource    = {dblp computer science bibliography, https://dblp.org}
}

@article{deepseek_r1,
  author       = {Daya Guo and
                  Dejian Yang and
                  Haowei Zhang and
                  Junxiao Song and
                  Peiyi Wang and
                  Qihao Zhu and
                  Runxin Xu and
                  Ruoyu Zhang and
                  Shirong Ma and
                  Xiao Bi and
                  Xiaokang Zhang and
                  Xingkai Yu and
                  Yu Wu and
                  Z. F. Wu and
                  Zhibin Gou and
                  Zhihong Shao and
                  Zhuoshu Li and
                  Ziyi Gao and
                  Aixin Liu and
                  Bing Xue and
                  Bingxuan Wang and
                  Bochao Wu and
                  Bei Feng and
                  Chengda Lu and
                  Chenggang Zhao and
                  Chengqi Deng and
                  Chong Ruan and
                  Damai Dai and
                  Deli Chen and
                  Dongjie Ji and
                  Erhang Li and
                  Fangyun Lin and
                  Fucong Dai and
                  Fuli Luo and
                  Guangbo Hao and
                  Guanting Chen and
                  Guowei Li and
                  Hao Zhang and
                  Hanwei Xu and
                  Honghui Ding and
                  Huazuo Gao and
                  Hui Qu and
                  Hui Li and
                  Jianzhong Guo and
                  Jiashi Li and
                  Jingchang Chen and
                  Jingyang Yuan and
                  Jinhao Tu and
                  Junjie Qiu and
                  Junlong Li and
                  J. L. Cai and
                  Jiaqi Ni and
                  Jian Liang and
                  Jin Chen and
                  Kai Dong and
                  Kai Hu and
                  Kaichao You and
                  Kaige Gao and
                  Kang Guan and
                  Kexin Huang and
                  Kuai Yu and
                  Lean Wang and
                  Lecong Zhang and
                  Liang Zhao and
                  Litong Wang and
                  Liyue Zhang and
                  Lei Xu and
                  Leyi Xia and
                  Mingchuan Zhang and
                  Minghua Zhang and
                  Minghui Tang and
                  Mingxu Zhou and
                  Meng Li and
                  Miaojun Wang and
                  Mingming Li and
                  Ning Tian and
                  Panpan Huang and
                  Peng Zhang and
                  Qiancheng Wang and
                  Qinyu Chen and
                  Qiushi Du and
                  Ruiqi Ge and
                  Ruisong Zhang and
                  Ruizhe Pan and
                  Runji Wang and
                  R. J. Chen and
                  R. L. Jin and
                  Ruyi Chen and
                  Shanghao Lu and
                  Shangyan Zhou and
                  Shanhuang Chen and
                  Shengfeng Ye and
                  Shiyu Wang and
                  Shuiping Yu and
                  Shunfeng Zhou and
                  Shuting Pan and
                  S. S. Li and
                  Shuang Zhou and
                  Shaoqing Wu and
                  Tao Yun and
                  Tian Pei and
                  Tianyu Sun and
                  Tao Wang and
                  Wangding Zeng and
                  Wen Liu and
                  Wenfeng Liang and
                  Wenjun Gao and
                  Wenqin Yu and
                  Wentao Zhang and
                  W. L. Xiao and
                  Wei An and
                  Xiaodong Liu and
                  Xiaohan Wang and
                  Xiaokang Chen and
                  Xiaotao Nie and
                  Xin Cheng and
                  Xin Liu and
                  Xin Xie and
                  Xingchao Liu and
                  Xinyu Yang and
                  Xinyuan Li and
                  Xuecheng Su and
                  Xuheng Lin and
                  X. Q. Li and
                  Xiangyue Jin and
                  Xiaojin Shen and
                  Xiaosha Chen and
                  Xiaowen Sun and
                  Xiaoxiang Wang and
                  Xinnan Song and
                  Xinyi Zhou and
                  Xianzu Wang and
                  Xinxia Shan and
                  Y. K. Li and
                  Y. Q. Wang and
                  Y. X. Wei and
                  Yang Zhang and
                  Yanhong Xu and
                  Yao Li and
                  Yao Zhao and
                  Yaofeng Sun and
                  Yaohui Wang and
                  Yi Yu and
                  Yichao Zhang and
                  Yifan Shi and
                  Yiliang Xiong and
                  Ying He and
                  Yishi Piao and
                  Yisong Wang and
                  Yixuan Tan and
                  Yiyang Ma and
                  Yiyuan Liu and
                  Yongqiang Guo and
                  Yuan Ou and
                  Yuduan Wang and
                  Yue Gong and
                  Yuheng Zou and
                  Yujia He and
                  Yunfan Xiong and
                  Yuxiang Luo and
                  Yuxiang You and
                  Yuxuan Liu and
                  Yuyang Zhou and
                  Y. X. Zhu and
                  Yanping Huang and
                  Yaohui Li and
                  Yi Zheng and
                  Yuchen Zhu and
                  Yunxian Ma and
                  Ying Tang and
                  Yukun Zha and
                  Yuting Yan and
                  Z. Z. Ren and
                  Zehui Ren and
                  Zhangli Sha and
                  Zhe Fu and
                  Zhean Xu and
                  Zhenda Xie and
                  Zhengyan Zhang and
                  Zhewen Hao and
                  Zhicheng Ma and
                  Zhigang Yan and
                  Zhiyu Wu and
                  Zihui Gu and
                  Zijia Zhu and
                  Zijun Liu and
                  Zilin Li and
                  Ziwei Xie and
                  Ziyang Song and
                  Zizheng Pan and
                  Zhen Huang and
                  Zhipeng Xu and
                  Zhongyu Zhang and
                  Zhen Zhang},
  title        = {DeepSeek-R1 incentivizes reasoning in LLMs through reinforcement learning},
  journal      = {Nat.},
  volume       = {645},
  number       = {8081},
  pages        = {633--638},
  year         = {2025},
  url          = {https://doi.org/10.1038/s41586-025-09422-z},
  doi          = {10.1038/S41586-025-09422-Z},
  bibsource    = {dblp computer science bibliography, https://dblp.org}
}

@inproceedings{VeRL,
  author       = {Guangming Sheng and
                  Chi Zhang and
                  Zilingfeng Ye and
                  Xibin Wu and
                  Wang Zhang and
                  Ru Zhang and
                  Yanghua Peng and
                  Haibin Lin and
                  Chuan Wu},
  title        = {HybridFlow: {A} Flexible and Efficient {RLHF} Framework},
  booktitle    = {Proceedings of the Twentieth European Conference on Computer Systems,
                  EuroSys 2025, Rotterdam, The Netherlands, 30 March 2025 - 3 April
                  2025},
  pages        = {1279--1297},
  publisher    = {{ACM}},
  year         = {2025},
  url          = {https://doi.org/10.1145/3689031.3696075},
  doi          = {10.1145/3689031.3696075},
  bibsource    = {dblp computer science bibliography, https://dblp.org}
}

@article{MathVision,
  author       = {Muhammad Awais Ahmad and
                  Tauqir Ahmed and
                  Muhammad Aslam and
                  Amjad Rehman and
                  Faten S. Alamri and
                  Saeed Ali Bahaj and
                  Tanzila Saba},
  title        = {MathVision: An Accessible Intelligent Agent for Visually Impaired
                  People to Understand Mathematical Equations},
  journal      = {{IEEE} Access},
  volume       = {13},
  pages        = {6155--6165},
  year         = {2025},
  url          = {https://doi.org/10.1109/ACCESS.2024.3514079},
  doi          = {10.1109/ACCESS.2024.3514079},
  bibsource    = {dblp computer science bibliography, https://dblp.org}
}

@inproceedings{MMMU,
  author       = {Xiang Yue and
                  Yuansheng Ni and
                  Tianyu Zheng and
                  Kai Zhang and
                  Ruoqi Liu and
                  Ge Zhang and
                  Samuel Stevens and
                  Dongfu Jiang and
                  Weiming Ren and
                  Yuxuan Sun and
                  Cong Wei and
                  Botao Yu and
                  Ruibin Yuan and
                  Renliang Sun and
                  Ming Yin and
                  Boyuan Zheng and
                  Zhenzhu Yang and
                  Yibo Liu and
                  Wenhao Huang and
                  Huan Sun and
                  Yu Su and
                  Wenhu Chen},
  title        = {{MMMU:} {A} Massive Multi-Discipline Multimodal Understanding and
                  Reasoning Benchmark for Expert {AGI}},
  booktitle    = {{IEEE/CVF} Conference on Computer Vision and Pattern Recognition,
                  {CVPR} 2024, Seattle, WA, USA, June 16-22, 2024},
  pages        = {9556--9567},
  publisher    = {{IEEE}},
  year         = {2024},
  url          = {https://doi.org/10.1109/CVPR52733.2024.00913},
  doi          = {10.1109/CVPR52733.2024.00913},
  bibsource    = {dblp computer science bibliography, https://dblp.org}
}

@inproceedings{MMStar,
  author       = {Lin Chen and
                  Jinsong Li and
                  Xiaoyi Dong and
                  Pan Zhang and
                  Yuhang Zang and
                  Zehui Chen and
                  Haodong Duan and
                  Jiaqi Wang and
                  Yu Qiao and
                  Dahua Lin and
                  Feng Zhao},
  editor       = {Amir Globersons and
                  Lester Mackey and
                  Danielle Belgrave and
                  Angela Fan and
                  Ulrich Paquet and
                  Jakub M. Tomczak and
                  Cheng Zhang},
  title        = {Are We on the Right Way for Evaluating Large Vision-Language Models?},
  booktitle    = {Advances in Neural Information Processing Systems 38: Annual Conference
                  on Neural Information Processing Systems 2024, NeurIPS 2024, Vancouver,
                  BC, Canada, December 10 - 15, 2024},
  year         = {2024},
  url          = {http://papers.nips.cc/paper\_files/paper/2024/hash/2f8ee6a3d766b426d2618e555b5aeb39-Abstract-Conference.html},
  bibsource    = {dblp computer science bibliography, https://dblp.org}
}

@misc{LMMs-Eval,
      title={LMMs-Eval: Reality Check on the Evaluation of Large Multimodal Models},
      author={Kaichen Zhang and Bo Li and Peiyuan Zhang and Fanyi Pu and Joshua Adrian Cahyono and Kairui Hu and Shuai Liu and Yuanhan Zhang and Jingkang Yang and Chunyuan Li and Ziwei Liu},
      year={2024},
      eprint={2407.12772},
      archivePrefix={arXiv},
      primaryClass={cs.CL},
      url={https://arxiv.org/abs/2407.12772},
}

@misc{bge-m3,
      title={BGE M3-Embedding: Multi-Lingual, Multi-Functionality, Multi-Granularity Text Embeddings Through Self-Knowledge Distillation}, 
      author={Jianlv Chen and Shitao Xiao and Peitian Zhang and Kun Luo and Defu Lian and Zheng Liu},
      year={2024},
      eprint={2402.03216},
      archivePrefix={arXiv},
      primaryClass={cs.CL}
}

@misc{dennis2020emergent,
      title={Emergent Complexity and Zero-shot Transfer via Unsupervised Environment Design}, 
      author={Michael Dennis and Natasha Jaques and Eugene Vinitsky and Alexandre Bayen and Stuart Russell and Andrew Critch and Sergey Levine},
      year={2021},
      eprint={2012.02096},
      archivePrefix={arXiv},
      primaryClass={cs.LG},
      url={https://arxiv.org/abs/2012.02096}, 
}

@misc{jiang2021prioritized,
      title={Prioritized Level Replay}, 
      author={Minqi Jiang and Edward Grefenstette and Tim Rocktäschel},
      year={2021},
      eprint={2010.03934},
      archivePrefix={arXiv},
      primaryClass={cs.LG},
      url={https://arxiv.org/abs/2010.03934}, 
}

@article{jiang2021replay,
  author       = {Minqi Jiang and
                  Michael Dennis and
                  Jack Parker{-}Holder and
                  Jakob N. Foerster and
                  Edward Grefenstette and
                  Tim Rockt{\"{a}}schel},
  title        = {Replay-Guided Adversarial Environment Design},
  journal      = {CoRR},
  volume       = {abs/2110.02439},
  year         = {2021},
  url          = {https://arxiv.org/abs/2110.02439},
  eprinttype   = {arXiv},
  eprint       = {2110.02439},
  bibsource    = {dblp computer science bibliography, https://dblp.org}
}

@misc{parkerholder2022evolving,
      title={Evolving Curricula with Regret-Based Environment Design}, 
      author={Jack Parker-Holder and Minqi Jiang and Michael Dennis and Mikayel Samvelyan and Jakob Foerster and Edward Grefenstette and Tim Rocktäschel},
      year={2023},
      eprint={2203.01302},
      archivePrefix={arXiv},
      primaryClass={cs.LG},
      url={https://arxiv.org/abs/2203.01302}, 
}

@misc{li2023generalization,
      title={Generalization through Diversity: Improving Unsupervised Environment Design}, 
      author={Wenjun Li and Pradeep Varakantham and Dexun Li},
      year={2023},
      eprint={2301.08025},
      archivePrefix={arXiv},
      primaryClass={cs.AI},
      url={https://arxiv.org/abs/2301.08025}, 
}

@misc{teoh2024improving,
      title={Improving Environment Novelty Quantification for Effective Unsupervised Environment Design}, 
      author={Jayden Teoh and Wenjun Li and Pradeep Varakantham},
      year={2025},
      eprint={2502.05726},
      archivePrefix={arXiv},
      primaryClass={cs.LG},
      url={https://arxiv.org/abs/2502.05726}, 
}

@misc{li2026agenticenvironmentengineeringlarge,
      title={Agentic Environment Engineering for Large Language Models: A Survey of Environment Modeling, Synthesis, Evaluation, and Application}, 
      author={Jiachun Li and Zhuoran Jin and Tianyi Men and Yupu Hao and Kejian Zhu and Lingshuai Wang and Dongqi Huang and Longxiang Wang and Shengjia Hua and Lu Wang and Jinshan Gao and Hongbang Yuan and Ruilin Xu and Kang Liu and Jun Zhao},
      year={2026},
      eprint={2606.12191},
      archivePrefix={arXiv},
      primaryClass={cs.CL},
      url={https://arxiv.org/abs/2606.12191}, 
}

@misc{men2025agentrewardbenchunifiedbenchmarkreward,
      title={Agent-RewardBench: Towards a Unified Benchmark for Reward Modeling across Perception, Planning, and Safety in Real-World Multimodal Agents}, 
      author={Tianyi Men and Zhuoran Jin and Pengfei Cao and Yubo Chen and Kang Liu and Jun Zhao},
      year={2025},
      eprint={2506.21252},
      archivePrefix={arXiv},
      primaryClass={cs.CL},
      url={https://arxiv.org/abs/2506.21252}, 
}

@misc{zhu2026mmrvwhatsleftunsaid,
      title={MMR-V: What's Left Unsaid? A Benchmark for Multimodal Deep Reasoning in Videos}, 
      author={Kejian Zhu and Zhuoran Jin and Hongbang Yuan and Jiachun Li and Shangqing Tu and Pengfei Cao and Yubo Chen and Kang Liu and Jun Zhao},
      year={2026},
      eprint={2506.04141},
      archivePrefix={arXiv},
      primaryClass={cs.CV},
      url={https://arxiv.org/abs/2506.04141}, 
}

@misc{li2026mmrlifepiecingreallifescenes,
      title={MMR-Life: Piecing Together Real-life Scenes for Multimodal Multi-image Reasoning}, 
      author={Jiachun Li and Shaoping Huang and Zhuoran Jin and Chenlong Zhang and Pengfei Cao and Yubo Chen and Kang Liu and Jun Zhao},
      year={2026},
      eprint={2603.02024},
      archivePrefix={arXiv},
      primaryClass={cs.CL},
      url={https://arxiv.org/abs/2603.02024}, 
}

@misc{jin2026looklightthinkheavy,
      title={Look Light, Think Heavy: What Multimodal Chain-of-Thought Reasoning Can and Cannot Do}, 
      author={Zhuoran Jin and Kejian Zhu and Hongbang Yuan and Yupu Hao and Pengfei Cao and Yubo Chen and Kang Liu and Jun Zhao},
      year={2026},
      eprint={2606.22565},
      archivePrefix={arXiv},
      primaryClass={cs.CL},
      url={https://arxiv.org/abs/2606.22565}, 
}

@misc{hao2026multisteptoolusereinforcementlearning,
      title={Why Multi-Step Tool-Use Reinforcement Learning Collapses and How Supervisory Signals Fix It}, 
      author={Yupu Hao and Zhuoran Jin and Huanxuan Liao and Kang Liu and Jun Zhao},
      year={2026},
      eprint={2606.26027},
      archivePrefix={arXiv},
      primaryClass={cs.CL},
      url={https://arxiv.org/abs/2606.26027}, 
}

@misc{hao2026pushinglimitsllmtool,
      title={Pushing the Limits of LLM Tool Calling via Experiential Knowledge Integration and Activation}, 
      author={Yupu Hao and Zhuoran Jin and Huanxuan Liao and Kang Liu and Jun Zhao},
      year={2026},
      eprint={2606.10875},
      archivePrefix={arXiv},
      primaryClass={cs.CL},
      url={https://arxiv.org/abs/2606.10875}, 
}

@misc{men2026physicsmultiturnlonghorizonplanning,
      title={The Physics of Multi-Turn Long-Horizon Planning: From Pre-training to Post-training via Single- and Multi-Teacher On-Policy Agentic Distillation}, 
      author={Tianyi Men and Zhuoran Jin and Kang Liu and Jun Zhao},
      year={2026},
      eprint={2607.24720},
      archivePrefix={arXiv},
      primaryClass={cs.CL},
      url={https://arxiv.org/abs/2607.24720}, 
}

@misc{jin2025omnirewardgeneralistomnimodalreward,
      title={Omni-Reward: Towards Generalist Omni-Modal Reward Modeling with Free-Form Preferences}, 
      author={Zhuoran Jin and Hongbang Yuan and Kejian Zhu and Jiachun Li and Pengfei Cao and Yubo Chen and Kang Liu and Jun Zhao},
      year={2025},
      eprint={2510.23451},
      archivePrefix={arXiv},
      primaryClass={cs.CL},
      url={https://arxiv.org/abs/2510.23451}, 
}

@misc{men2026empoweringguiagentsautonomous,
      title={Empowering GUI Agents via Autonomous Experience Exploration and Hindsight Experience Utilization for Task Planning}, 
      author={Tianyi Men and Zhuoran Jin and Pengfei Cao and Yubo Chen and Kang Liu and Jun Zhao},
      year={2026},
      eprint={2606.27330},
      archivePrefix={arXiv},
      primaryClass={cs.CL},
      url={https://arxiv.org/abs/2606.27330}, 
}

@misc{li2026monocularavatarreconstructioncascaded,
      title={Monocular Avatar Reconstruction via Cascaded Diffusion Priors and UV-Space Differentiable Shading}, 
      author={Hong Li and Minqi Meng and Yanjun Liang and Chongjie Ye and Houyuan Chen and Weiqing Xiao and Xianda Guo and Guojun Lei and Xuhui Liu and Chaojie Yang and Yanlun Peng and Hao Zhao and Baochang Zhang},
      year={2026},
      eprint={2606.28144},
      archivePrefix={arXiv},
      primaryClass={cs.CV},
      url={https://arxiv.org/abs/2606.28144}, 
}

\appendix

\section{More Related Works}
\label{appendix:more_related_works}

We discuss the works most closely related to our main thread in Section~\ref{sec:related_works}. 
In this appendix, we provide a detailed review of several related areas about environment~\cite{li2026agenticenvironmentengineeringlarge} and agent learning~\cite{men2026physicsmultiturnlonghorizonplanning,hao2026multisteptoolusereinforcementlearning,hao2026pushinglimitsllmtool}.

\subsection{Multimodal Environments}

Recent research on multimodal agents~\cite{men2025agentrewardbenchunifiedbenchmarkreward} has gradually shifted from static datasets~\cite{zhu2026mmrvwhatsleftunsaid, li2026mmrlifepiecingreallifescenes,jin2025omnirewardgeneralistomnimodalreward, li2026monocularavatarreconstructioncascaded} to dynamic interactive environments~\cite{SpatialEvo}. Existing studies extend multimodal environments from different perspectives. Some works focus on scalable construction of Web and GUI environments~\cite{InfiniteWeb, Gym-Anything, men2026empoweringguiagentsautonomous}. Other works further use game environments for reinforcement learning of VLM agents, emphasizing verifiable feedback, long-horizon interaction, and world-model reasoning~\cite{Game-RL,VAGEN}. Meanwhile, some works mainly focus on evaluation. Several studies build standardized benchmarks for multimodal game agents~\cite{GameWorld, PokeGym, ARC-AGI-3}. Most related to our work are more general multimodal environment systems such as Gym-V and VisGym, which unify diverse visual tasks into multi-turn interaction frameworks and provide a foundation for studying multimodal environment scaling.

\subsection{Sample-Level Environment Quality}

Another line of work examines the quality of constructed environments, typically focusing on whether each individual environment is executable and provides reliable reward signals~\cite{SWE-smith,R2E-Gym}. Common practices include using sandbox execution and unit tests to verify whether a task can run correctly~\cite{SWE-Gym,V-GameGym}. Many works further validate the correctness of state transitions and task solvability with golden trajectories or expert tool sequences~\cite{Agent2World, AutoForge}. Similarly, for multimodal environments where images or videos are difficult to test directly with unit tests, metric such as interaction consistency is often used as a correctness check~\cite{DreamGen,Matrix-game}. GUI and Web environments also commonly combine execution-based evaluators with expert review~\cite{AutoWebWorld,InterCode}. These methods provide important guarantees for the executability, solvability, and evaluation reliability of individual environments. However, they mainly answer whether a single environment is valid. In contrast, our work further studies the distribution-level quality of an environment pool when it is used as a training distribution.

\subsection{Environment Diversity}

A key aspect of distribution-level quality is diversity~\cite{SWE-World}. Existing works usually control diversity in two ways. The first line removes semantically similar tasks using information such as embedding similarity~\cite{Agent_World_Model}. For example some works reduce repeated samples through embedding-based deduplication, topic dispersion analysis, or clustering of game seeds~\cite{Agent_World_Model,EnvScaler,V-GameGym}. The second line controls diversity through structured construction. Some works analyze diversity from the perspectives of tool categories and tool combinations~\cite{Api-bank,AutoForge}, while Web and GUI environments often judge by the coverage of application scenarios~\cite{Mind2web,WebArena,Mcp-universe}. However, these methods mostly measure surface-level diversity, such as task descriptions or domain sources, and may not reflect the diversity of abilities that the model actually learns during training. In contrast, our work studies the diversity of underlying logic and dynamics of the environment through agent behavior and gradient-level mechanism analysis.

\subsection{Difficulty Grading and Curriculum Learning}

Difficulty grading and scheduling within an environment set are also crucial to training-set quality. Existing works often characterize task complexity using structural parameters, such as the number of tools, state entities, or code scale~\cite{OSWorld-MCP,LOGIGEN}. Some studies calibrate difficulty with strong models or human experts~\cite{gg-bench,AI_GAMESTORE}. For long-horizon planning tasks, difficulty is also adjusted by planning horizon or the number of search iterations~\cite{MobileDreamer, VLWM}.
Several works further use difficulty schedules for curriculum learning. Early curriculum learning methods often rely on predefined stages or static easy-to-hard ordering~\cite{e2h_reasoner}. Recent RL methods for LLMs further emphasize the dynamic nature of difficulty. RLVE~\cite{RLVE} adaptively adjusts the difficulty of problems in verifiable environments based on model performance. ADCL~\cite{ADCL} periodically re-estimates sample difficulty to mitigate difficulty shift during training. VCRL~\cite{Vcrl} selects samples with strong learning signals for the current model using group reward variance for rollouts. AdaCuRL~\cite{Adacurl} performs curriculum learning with coarse-to-fine difficulty estimation. WebRL~\cite{webrl} constructs an online self-evolving curriculum by generating tasks from failed trajectories.
Existing difficulty grading for multimodal environments is also often based on environment scale~\cite{visgym, gym-v}. However, the difficulty of multimodal environments does not only come from task scale, but also from multimodal-specific ability bottlenecks. Prior works provide limited systematic analysis of difficulty grading and curriculum learning tailored to these multimodal-specific factors.

\subsection{Unsupervised Environment Design}
\label{app:ued}

Unsupervised Environment Design (UED) automatically generates or
selects environments according to the current learning state
of an agent, thereby inducing an adaptive curriculum. PAIRED
\citep{dennis2020emergent} formulates environment design as an
adversarial minimax-regret problem and generates environments near the
current capability frontier. Subsequent work improves how
environments are reused. PLR
\citep{jiang2021prioritized} prioritizes previously encountered levels
according to their estimated learning potential. REPAIRED
\citep{jiang2021replay} combines adversarial environment generation
with prioritized replay, while ACCEL
\citep{parkerholder2022evolving} edits and mutates previously generated
levels to produce challenging environments. Beyond
learning potential, Diversity-UED \citep{li2023generalization} observes
that UED may repeatedly select similar environments and therefore
introduces an explicit environment-distance measure to promote
diversity. CENIE \citep{teoh2024improving} further quantifies
curriculum-aware environment novelty using the student's historical
state-action coverage. Unlike UED, which typically generates instances online, AES and HDC select and organize a fixed, heterogeneous pool of multimodal environment types. The two settings therefore address different stages of environment design.

\section{Experiment Details}
\label{appendix:exp_details}

This section describes the experimental settings in detail. We first introduce the settings shared by all experiments, and then describe the specific settings for each group of experiments.

\subsection{General Settings}
\label{appendix:general_settings}

\textbf{Training.}
We use Qwen3-VL-4B-Instruct~\cite{Qwen3} as the base model for training. In the main experiments, we further include Qwen3-VL-8B-Instruct to evaluate whether our method remains effective with a larger model. To ensure a fair comparison, we follow the setting of RLVE~\cite{RLVE} and keep the compute budget the same across different training runs. Specifically, all runs use the same total number of training samples, which is set to 7,680 in our experiments. We use GRPO as the training algorithm~\cite{deepseek_r1}, and set the rollout group size for each prompt to 16. Our reinforcement learning framework is modified from VeRL~\cite{VeRL} and follow its default implementation. We thank the authors for their excellent open source work. All training experiments are conducted on 4 A100 GPUs (80G).

\textbf{Evaluation.}
The main evaluation in this paper is conducted on multimodal environments, with the goal of measuring both model performance and generalization. We consider two evaluation settings: in-distribution (ID) and out-of-distribution (OOD). ID evaluation uses unseen instances generated from environment types that appear during training. The specific set of ID environments may vary across experiments. OOD evaluation uses 30 held-out environments randomly selected from our pool of 200 environments. These 30 environments are never used in any training experiment in this paper. Therefore, all OOD results refer to evaluation on this same held-out set. For each environment type, we generate 100 instances for evaluation. Our evaluation scripts for multimodal environments are based on the implementation of Gym-V~\cite{gym-v}. We thank the authors for their excellent open source work.


\subsection{Settings for Preliminary Experiments}
\label{appendix:preliminary_details}
For the environment scaling experiment in Section~\ref{sec:scaling_failure}, we conduct all experiments on Qwen3-VL-4B and keep the compute budget fixed. The environment sets are nested across different scales, meaning that a larger set contains the smaller set as a subset. In this experiment, the ID evaluation set is defined as the smallest subset of 20 environments, so that all trained models are evaluated on the same group of ID environments. For the gradient conflict analysis in Section~\ref{sec:negative_transfer}, we sample 10 batches from each environment. Each batch contains 10 samples, and each prompt uses 16 rollouts. We then compute the gradient direction for each batch. For the error analysis in Section~\ref{sec:bottleneck}, two human annotators independently label the model errors. The annotations are then cross checked, and disagreements are resolved to obtain the final labels.

\subsection{Settings for Main Experiments}

\textbf{Training.}
In the main experiments, we further include Qwen3-VL-8B-Instruct to test whether our method is still effective at a larger scale. We use the same compute budget for all methods. Note that the All Envs setting trains on the pool of 170 environments, after excluding the 30 held-out environments. This setting is used to analyze whether naive scaling of the environment pool is effective. It should also be noted that among the 30 environments in the AES-selected set, there are 22 single-turn and 8 multi-turn environments. Among the 30 OOD environments in the held-out set, there are 20 single-turn and 10 multi-turn environments.

\textbf{Evaluation.}
For the main environment evaluation, ID has slightly different meanings for different methods. For all settings except Random-K, ID refers to the 30 environments selected by AES in Section~\ref{sec:bcrc}. For Random-K, ID refers to the 30 randomly selected training environments. In addition, we evaluate the trained models on common general multimodal benchmarks to examine whether training on interactive environments harms their general perception and reasoning ability. We use three benchmarks of different types: MathVision~\cite{MathVision}, MMMU~\cite{MMMU} with the validation split, and MMStar~\cite{MMStar}. The evaluation scripts for these general multimodal benchmarks are based on lmms-eval~\cite{LMMs-Eval}. We thank the authors for their excellent open source implementation.

\section{Details of Ability-aware Environment Selection (AES)}
\label{appendix:diversity_selection_details}

\subsection{Atomic Ability Segmentation}

We first use Qwen3-VL-4B and Gemini-3-Flash to generate 20 trajectories for each environment. In practice, we sample 5 instances for each environment type, and generate 4 trajectories for each instance with each model. For atomic ability segmentation, we use GPT-5 as the segmentation model. The prompt is shown in Appendix~\ref{appendix:atomic_ability_prompt}. The goal is to decompose a complete agent trajectory into reusable and interpretable behavior units. When defining the segmentation granularity, we first rely on the model's understanding, and then manually inspect the results to slightly adjust the granularity.

\subsection{Construction of Environment Meta Ability Profiles}

After obtaining atomic ability segmentations from 40 trajectories for each environment, we merge these atomic abilities into environment level meta abilities. The resulting meta ability profile is used to represent the underlying logic of environment.

During merging, we first use GPT-5 to combine abilities with the same meaning but slightly different wording. We then filter out atomic abilities whose frequency is lower than 10\%, treating them as noise. For the remaining abilities, we use GPT-5 to classify them into core abilities and soft abilities. Core abilities refer to abilities that are essential for solving the task, such as path planning, while soft abilities refer to auxiliary abilities, such as answer formatting. These two types of abilities are assigned different weights in later diversity based selection~\ref{sec:bcrc}. We also record transition edges between abilities as ability graph information. Finally, we aggregate core abilities, soft abilities, ability graph edges, and their frequencies into the meta ability profile of each environment.

\subsection{Settings of the AES Method}

Our selection method computes the utility of each candidate environment based on three factors: covering as many core abilities as possible, reducing redundancy, and reducing conflict. We set the weights of these three terms, denoted by $\lambda$, to 1 in our current experiments.
For redundancy, we use the maximum profile similarity between the candidate environment and the already selected environments as the redundancy penalty. To compute cosine similarity, we embed core abilities, soft abilities, and ability graph edges. These three components are weighted differently. The weight of a core ability vector is its frequency, while the weights of soft ability vectors and edge vectors are $0.3$ times their frequencies. This is because soft abilities should not dominate the redundancy estimate. For example, two completely different environments may both require the ability of answer formatting. Finally, we measure conflict using gradients, following the same implementation as described in Appendix~\ref{appendix:preliminary_details}.
Table~\ref{tab:selected_envs_desc} lists the 30 high quality environments selected by AES. These environments are selected to achieve broad coverage, low redundancy, and low conflict. The table also provides a brief task description for each environment.

\clearpage
\onecolumn

\begin{longtable}{p{0.24\textwidth} p{0.70\textwidth}}
\toprule
\textbf{Environment} & \textbf{Task Description} \\
\midrule
\endfirsthead

\multicolumn{2}{l}{\textit{Table~\ref{tab:selected_envs_desc} continued from previous page.}} \\
\toprule
\textbf{Environment} & \textbf{Task Description} \\
\midrule
\endhead

\midrule
\multicolumn{2}{r}{\textit{Continued on next page}} \\
\endfoot

\bottomrule
\caption{Selected environments and task descriptions.}
\label{tab:selected_envs_desc}
\\
\endlastfoot

DoorKey &
In a first-person partial-view room, find the key, unlock the door, pass through, and reach the green goal. \\

Tents-QA &
Visual QA on Tents and Trees: place tents next to trees while satisfying row/column counts and non-adjacency constraints. \\

PegJump &
Jump one peg over another to remove it; goal is to finish with exactly one peg remaining. \\

Arc1D &
Infer the 1D input-output transformation rule from examples and predict the output for the test input. \\

LavaGap &
Cross a lava gap and reach the goal in first-person view without stepping on lava. \\

Minimum Spanning Tree Counting &
Count the number of distinct minimum spanning trees in a weighted undirected graph, modulo a given value. \\

Othello &
Two-player Othello: place discs to flip opponent pieces; win by having more pieces when the game ends. \\

Frozen Lake &
Navigate from start to goal on a slippery frozen-lake grid without falling into holes. \\

Multi Room &
Explore multiple connected rooms in first-person view; use keys and doors to reach the goal. \\

Mini Sudoku &
Fill a Mini Sudoku so each row, column, and block contains each digit exactly once. \\

Tree Even Partitioning &
Partition a tree with $N\times K$ vertices into $N$ connected groups of $K$ vertices each. \\

Sum Triangle Area &
Compute the sum of areas of all triangles formed by given 2D points, where collinear triples count as 0. \\

Matchstick Equation &
Fix a broken matchstick equation by moving one match per action until the equation is mathematically correct. \\

Knight Swap &
Determine whether white and black knights can swap positions via legal L-moves; output the move sequence or ``impossible''. \\

Texas Holdem &
Two-player limit Texas Hold'em: form the best 5-card hand from hole and community cards; win via call, raise, fold, or check. \\

Hue-QA &
Visual QA on a color-gradient board about cell colors, gradient patterns, and color matching. \\

Circuit Logic &
Given input assignments and a logic-circuit diagram, compute the circuit output, either 0 or 1. \\

Weighted Binarytree &
Build a binary tree with fixed inorder traversal $0,\ldots,N-1$ that maximizes the given recursive score function. \\

Nine Puzzle &
Transform the start grid into the target grid using bounded cyclic row/column shifts. \\

Math Path &
Find a path from S to G on a number grid whose visited cells sum to the target value. \\

Minimum Chromatic Number &
Find the minimum number of colors needed to properly color the vertices of an undirected graph. \\

Tree To Traversal &
Read a binary tree visualization and output preorder, inorder, and postorder traversals. \\

Chess Ranger-QA &
Visual QA on Chess Ranger: capture-only chess puzzles aiming to leave exactly one piece on the board. \\

Klo Blocks &
Redistribute values via adjacent $\pm 1$ operations to maximize the longest contiguous subarray with all values at least $K$. \\

Rush Hour &
Slide cars to free the red car [X] and drive it out through the right edge. \\

Largest Island &
Find the maximum area of a 4-connected land island in a binary grid; return 0 if none exists. \\

Grid Local Minimum Counting &
Count valid numberings of 1 to $N\times M$ such that marked cells are exactly the 8-neighbor local minima. \\

Binary Matrix &
For each cell in a binary matrix, compute the Manhattan distance to the nearest 0. \\

Ultra TicTacToe-QA &
Visual QA on Ultimate Tic-Tac-Toe, consisting of nine nested 3$\times$3 boards, about rules, moves, or scoring. \\

Ska Rock Garden &
Choose coordinate swaps on points to minimize the axis-aligned bounding-box perimeter, with total swap cost used as a tie-breaker. \\

\end{longtable}

\clearpage
\twocolumn

\section{Details of Hierarchical Difficulty Curriculum (HDC)}
\label{app:curriculum_details}

\subsection{Pseudocode of HDC}
\label{appendix:Pseudocode_of_Curriculum_Learning}
This section provides the pseudocode of the Nested Harness-Annealed Curriculum in Algorithm~\ref{alg:nested_curriculum} to facilitate understanding of our curriculum design. The core idea is that each environment maintains its own curriculum state, including the current harness frontier $r_e$ and the state-scale difficulty frontier $u_e$. During training, we first sample an environment, then sample a harness level and a scale level, and finally generate a concrete training instance.

\begin{algorithm}[t]
\caption{Hierarchical Difficulty Curriculum}
\label{alg:nested_curriculum}
\begin{algorithmic}[1]
\Require Training environments $\mathcal{E}_{\mathrm{train}}$
\Require Harness sample distribution $D_e$ and scale range $[\ell_e, u_e^\star]$, $\ell_e = 0$ 
\Require Thresholds $\tau_{\mathrm{scale}}$ and $\tau_{\mathrm{harness}}$
\State Initialize $r_e=0$ and $u_e=\ell_e$ for each environment $e \in \mathcal{E}_{\mathrm{train}}$
\For{each training step}
    \State Sample $e \sim \mathcal{E}_{\mathrm{train}}$
    \State Sample $h \sim D_e(h \mid r_e)$
    \State Sample $s \sim \mathrm{Uniform}(\{\ell_e,\ldots,u_e\})$
    \State Generate an instance with $(e,h,s)$
    \State Update the policy with RL
    \State Estimate recent performance $p_e$ on scale $u_e$
    \If{$p_e > \tau_{\mathrm{scale}}$}
        \State $u_e \leftarrow \min(u_e+1, u_e^\star)$
    \EndIf
    \If{$u_e = u_e^\star$ and $p_e > \tau_{\mathrm{harness}}$}
        \State $r_e \leftarrow r_e + 1$; reset $u_e \leftarrow \ell_e$
    \EndIf
\EndFor
\end{algorithmic}
\end{algorithm}

As described in Section \ref{sec:nested_curriculum}, this curriculum is nested. Within each harness level, the model first progresses along the state-scale difficulty axis. Once the target scale is reached, we weaken the harness, advance the outer curriculum frontier, and restart the growth of state-scale difficulty.

\subsection{Harness Level}

This section presents detailed summarization of the harness level discussed in Section~\ref{sec:harness_curriculum}. Here, $H_0$ represents the strongest harness, where the model can simultaneously access textual observations, text state, text hints, and complete rule descriptions. As the level increases, auxiliary information is gradually removed. Finally, $H_4$ retains only basic information such as visual observations and task descriptions.

\begin{table}[t]
\centering
\small
\resizebox{\columnwidth}{!}{
\begin{tabular}{c|cccc|c}
\toprule
Level & Text Obs. & Text State & Text Hint & Rule & Basic Info. \\
\midrule
$H_0$ & \checkmark & \checkmark & \checkmark & \checkmark & \checkmark \\
$H_1$ & -- & \checkmark & \checkmark & \checkmark & \checkmark \\
$H_2$ & -- & -- & \checkmark & \checkmark & \checkmark \\
$H_3$ & -- & -- & -- & \checkmark & \checkmark \\
$H_4$ & -- & -- & -- & -- & \checkmark \\
\bottomrule
\end{tabular}
}
\caption{
Harness levels used in the outer curriculum.}
\label{tab:harness_levels}
\end{table}


\subsection{State-Scale Implementation Details}

This section describes the implementation details of the inner curriculum. In our method, the inner curriculum progresses according to state-scale difficulty levels, such as graph size, grid size, obstacle density, or the number of entities. Since different environments support different state-scale parameters, Table~\ref{tab:scale_parameters} summarizes the adjustable state-scale factors for the 30 selected environments. The progression of the inner curriculum follows the implementation strategy of RLVE~\cite{RLVE}.
\clearpage
\onecolumn

\begin{longtable}{
>{\raggedright\arraybackslash}p{0.20\textwidth}
>{\centering\arraybackslash}p{0.12\textwidth}
>{\raggedright\arraybackslash}p{0.62\textwidth}
}
\toprule
\textbf{Environment} & \textbf{Supported} & \textbf{Scale Parameters} \\
\midrule
\endfirsthead

\multicolumn{3}{l}{\textit{Table~\ref{tab:scale_parameters} continued from previous page.}} \\
\toprule
\textbf{Environment} & \textbf{Supported} & \textbf{Scale Parameters} \\
\midrule
\endhead

\midrule
\multicolumn{3}{r}{\textit{Continued on next page}} \\
\endfoot

\bottomrule
\caption{Scale controllability of selected environments.}
\label{tab:scale_parameters}
\\
\endlastfoot

DoorKey &
Yes &
\texttt{size}: map size; e.g., \texttt{size=5/6/8/16}. \\

Tents-QA &
Yes &
\texttt{grid\_size}; \texttt{num\_trees}; e.g., \texttt{grid\_size=[8,8]}, \texttt{num\_trees=12}. \\

PegJump &
No &
-- \\

Arc1D &
Yes &
\texttt{min\_size/max\_size}: sequence length; \texttt{num\_train}: number of examples; e.g., \texttt{min\_size=10}, \texttt{max\_size=30}, \texttt{num\_train=3}. \\

LavaGap &
Yes &
\texttt{size}: map size; e.g., \texttt{size=5/6/7}. \\

Minimum Spanning Tree Counting &
Yes &
\texttt{max\_n}: number of graph nodes; \texttt{edge\_ratio}: edge density; e.g., \texttt{max\_n=14}, \texttt{edge\_ratio=2.5}. \\

Othello &
Yes &
\texttt{board\_size}: board side length; e.g., \texttt{board\_size=4/6/8/10}. \\

Frozen Lake &
Yes &
\texttt{size}: grid size; \texttt{num\_holes}; e.g., \texttt{size=4}, \texttt{num\_holes=5}. \\

MultiRoom &
Yes &
\texttt{min\_num\_rooms/max\_num\_rooms}: room-count range; \texttt{max\_room\_size}: maximum room size; e.g., \texttt{min\_num\_rooms=2}, \texttt{max\_num\_rooms=6}, \texttt{max\_room\_size=10}. \\

Mini Sudoku &
Yes &
\texttt{min\_empty/max\_empty}: number of blank cells in a fixed 4$\times$4 grid; e.g., \texttt{min\_empty=8}, \texttt{max\_empty=12}. \\

Tree Even Partitioning &
Yes &
\texttt{max\_n}: number of tree groups; \texttt{max\_k}: group size; e.g., \texttt{max\_n=10}, \texttt{max\_k=6}. \\

Sum Triangle Area &
Yes &
\texttt{max\_n}: number of 2D points; e.g., \texttt{max\_n=9}. \\

Matchstick Equation &
Yes &
\texttt{break\_moves}: number of perturbation moves; \texttt{enforce\_min\_distance}: minimum repair-distance constraint; e.g., \texttt{break\_moves=2}, \texttt{enforce\_min\_distance=True}. \\

Knight Swap &
Yes &
\texttt{min\_nodes/max\_nodes}: board graph size; \texttt{min\_steps/max\_steps}: solution depth; e.g., \texttt{min\_nodes=9}, \texttt{max\_nodes=10}, \texttt{min\_steps=8}, \texttt{max\_steps=20}. \\

Texas Holdem &
No &
-- \\

Hue-QA &
Yes &
\texttt{board\_size}: color-board size; \texttt{num\_lines}: number of gradient lines; e.g., \texttt{board\_size=8}, \texttt{num\_lines=6}. \\

Circuit Logic &
Yes &
\texttt{min\_terms/max\_terms}: number of logic terms; \texttt{min\_inputs/max\_inputs}: number of inputs; e.g., \texttt{min\_inputs=4}, \texttt{max\_inputs=6}, \texttt{min\_terms=4}, \texttt{max\_terms=8}. \\

Weighted Binarytree &
Yes &
\texttt{max\_n}: number of tree nodes; \texttt{max\_score}: weight range; e.g., \texttt{max\_n=12}, \texttt{max\_score=20}. \\

Nine Puzzle &
Yes &
\texttt{max\_n\_m}: grid size; \texttt{steps}: number of scrambling steps; e.g., \texttt{max\_n\_m=4}, \texttt{steps=12}. \\

Math Path &
Yes &
\texttt{size}: grid size; e.g., \texttt{size=4}. \\

Minimum Chromatic Number &
Yes &
\texttt{max\_n}: number of graph nodes; \texttt{edge\_density}: graph density; e.g., \texttt{max\_n=14}, \texttt{edge\_density=0.6}. \\

Tree To Traversal &
Yes &
\texttt{min\_nodes/max\_nodes}: binary-tree node-count range; e.g., \texttt{min\_nodes=12}, \texttt{max\_nodes=16}. \\

Chess Ranger-QA &
Yes &
\texttt{num\_pieces}: number of chess pieces on the board; e.g., \texttt{num\_pieces=8}. \\

Klo Blocks &
Yes &
\texttt{N}: array or board length; e.g., \texttt{N=7}. \\

Rush Hour &
Yes &
\texttt{difficulty}: predefined difficulty level; e.g., \texttt{difficulty=easy/medium/hard}. \\

Largest Island &
Yes &
\texttt{min\_num\_islands/max\_num\_islands}: number of islands; \texttt{max\_island\_size}: maximum island size; e.g., \texttt{min\_num\_islands=4}, \texttt{max\_num\_islands=8}, \texttt{max\_island\_size=20}. \\

Grid Local Minimum Counting &
Yes &
\texttt{max\_n\_m}: maximum grid side length; e.g., \texttt{max\_n\_m=5}. \\

Binary Matrix &
Yes &
\texttt{min\_n/max\_n}: matrix size; e.g., \texttt{min\_n=5}, \texttt{max\_n=8}. \\

Ultra TicTacToe-QA &
No &
-- \\

SkaRock Garden &
Yes &
\texttt{max\_n}: number of points; e.g., \texttt{max\_n=7}. \\

\end{longtable}

\clearpage
\twocolumn

\section{More Experimental Results}
\label{appendix:more_sxp_results}

\begin{table}[t]
\centering
\small
\setlength{\tabcolsep}{5.5pt}
\renewcommand{\arraystretch}{1.02}
\begin{tabular}{@{}lccc@{}}
\toprule
\textbf{Method} & \textbf{OOD-ST} & \textbf{OOD-MT} & \textbf{Rel.} \\
\midrule
Base Model              
& 13.8 & 9.5  & 0.0  \\

Description Embedding   
& 16.3 & 8.3  & 2.8  \\

Code Embedding          
& 18.6 & 9.9  & 19.5 \\

\rowcolor{gray!10}
\textbf{\textcolor{myred}{AES}}                     
& \textbf{21.0} & \textbf{12.2} & \textbf{40.3} \\
\bottomrule
\end{tabular}
\caption{
Comparison with representation-based environment selection.
Description Embedding selects environments based on embeddings of textual environment descriptions, while Code Embedding selects environments based on embeddings of key environment code.
}
\label{tab:surface_selection}
\end{table}

\subsection{Expanded-Budget Experiments}
\label{app:expanded_budget}

Our main experiments use a fixed total training budget to ensure
compute-matched comparisons across different environment-set sizes in Section\ref{sec:preliminaries}.
However, under this setting, increasing the number of environments
reduces the number of training samples allocated to each environment.
To control for this potential confounding factor, we conduct additional
expanded-budget experiments, where each environment is assigned 256
training samples and the total budget grows linearly with the number of
training environments.

\begin{figure}[t]
    \centering
    \includegraphics[width=\linewidth]
    {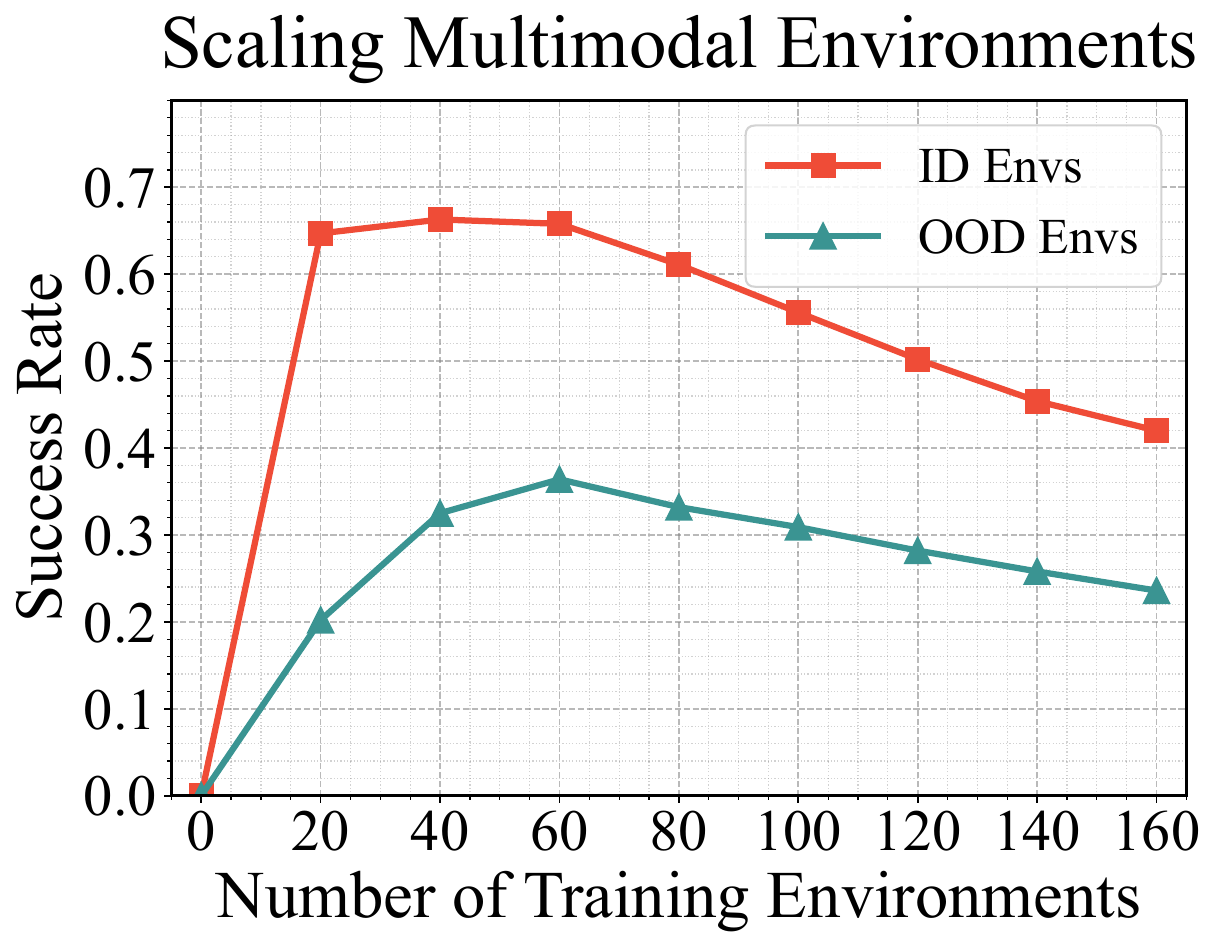}
    \caption{
    Multimodal environment scaling under an expanded training budget.
    We fix the number of training samples per environment to 256.
    }
    \label{fig:expanded_budget_scaling}
\end{figure}

As shown in Figure~\ref{fig:expanded_budget_scaling}, performance
remains non-monotonic even when the total training budget is expanded.
The ID success rate peaks at 66.3\% with 40 environments and decreases
to 42.1\% with 160 environments. Similarly, the OOD success rate peaks
at 36.4\% with 60 environments and decreases to 23.6\% with 160
environments. Therefore, the scaling degradation observed in
Section~\ref{sec:scaling_failure} cannot be explained solely by insufficient
per-environment training samples.

We further compare training on all environments with AES under the
same allocation of 256 samples per environment. As shown in
Table~\ref{tab:expanded_budget_main}, despite using approximately 5.7 times more data, All Envs performs worse than AES across all evaluation settings. These results further demonstrate that the
improvements of AES arise from constructing a more effective training
distribution rather than from the fixed-budget setting.

\begin{table}[t]
    \centering
    \small
    \setlength{\tabcolsep}{6pt}
    \begin{tabular}{lrcccc}
        \toprule
        \multirow{2}{*}{Method}
        & \multirow{2}{*}{\# Envs.}
        & \multicolumn{2}{c}{ID Envs.}
        & \multicolumn{2}{c}{OOD Envs.} \\
        \cmidrule(lr){3-4}
        \cmidrule(lr){5-6}
        & & ST & MT & ST & MT \\
        \midrule
        Base Model
        & --   & 13.1 & 11.9 & 13.8 & 9.5  \\

        All Envs.
        & 170  & 33.1 & 20.6 & 20.5 & 9.8  \\

        \textbf{AES}
        & \textbf{30} 
        & \textbf{37.7} & \textbf{25.4}
        & \textbf{21.0} & \textbf{12.2} \\
        \bottomrule
    \end{tabular}
    \caption{
    Main results under the expanded-budget setting. Each training
    environment contributes 256 samples. ST denotes single-turn
    success rate and MT denotes normalized return in multi-turn
    environments.
    }
    \label{tab:expanded_budget_main}
\end{table}

\subsection{Performance Transfer across Modalities}
\label{appendix:performance_transfer_across_modalities}

In the preliminary experiments \ref{sec:negative_transfer}, we analyze performance transfer across environments under both text-only and multimodal settings. Due to space limitations, the main text only reports a summarized table. Here, we provide the complete performance transfer matrices in Table~\ref{tab:text_transfer_matrix} and Table~\ref{tab:mm_transfer_matrix}.

\begin{table*}[t]
\centering
\small
\setlength{\tabcolsep}{4.2pt}
\begin{tabular}{lcccccc}
\toprule
\multirow{2}{*}{\textbf{Train env.}} 
& \multicolumn{6}{c}{\textbf{Text-symbolic evaluation environment Acc. (\%)}} \\
\cmidrule(lr){2-7}
& addition\_table & mini\_sudoku & rectangle\_count & rotate\_matrix & spiral\_matrix & tictactoe\_qa \\
\midrule
Base Model
& 16.5 & 55.2 & 11.0 & 72.5 & 76.2 & 47.6 \\
\midrule
addition\_table 
& \textbf{37.5} \poschg{21.0} & 59.3 \poschg{4.1} & 18.5 \poschg{7.5} & 79.5 \poschg{7.0} & 83.2 \poschg{7.0} & 48.1 \poschg{0.5} \\
mini\_sudoku   
& 24.6 \poschg{8.1} & \textbf{82.1} \poschg{26.9} & 19.3 \poschg{8.3} & 81.3 \poschg{8.8} & 80.1 \poschg{3.9} & 45.3 \negchg{2.3} \\
rectangle\_count 
& 17.6 \poschg{1.1} & 55.0 \negchg{0.2} & \textbf{61.0} \poschg{50.0} & 77.1 \poschg{4.6} & 79.6 \poschg{3.4} & 42.1 \negchg{5.5} \\
rotate\_matrix 
& 21.5 \poschg{5.0} & 56.0 \poschg{0.8} & 21.0 \poschg{10.0} & \textbf{92.0} \poschg{19.5} & 93.0 \poschg{16.8} & 43.5 \negchg{4.1} \\
spiral\_matrix 
& 22.1 \poschg{5.6} & 56.7 \poschg{1.5} & 19.8 \poschg{8.8} & 92.8 \poschg{20.3} & \textbf{93.7} \poschg{17.5} & 42.1 \negchg{5.5} \\
tictactoe\_qa  
& 18.8 \poschg{2.3} & 54.5 \negchg{0.7} & 13.5 \poschg{2.5} & 75.7 \poschg{3.2} & 82.2 \poschg{6.0} & \textbf{55.8} \poschg{8.2} \\
\midrule
mixed6         
& 36.1 \poschg{19.6} & 80.2 \poschg{25.0} & 60.0 \poschg{49.0} & 93.3 \poschg{20.8} & 94.9 \poschg{18.7} & 52.6 \poschg{5.0} \\
$\Delta$      
& -1.4 & -1.9 & -1.0 & +1.3 & +1.2 & -3.2 \\
\bottomrule
\end{tabular}
\caption{
Transfer matrix in the text-symbolic setting.
Each row denotes the training environment, and each column denotes the evaluation environment.
All values are reported as accuracy in percentage. Colored values in parentheses denote the absolute accuracy change relative to the Base Model on the same evaluation environment.
Diagonal entries correspond to single-environment training performance.
The last row reports the change from mixed training to the corresponding single-environment result,
i.e., $\Delta_j = M_j - T_{j,j}$.
}
\label{tab:text_transfer_matrix}
\end{table*}

\begin{table*}[t]
\centering
\small
\setlength{\tabcolsep}{4.2pt}
\begin{tabular}{lcccccc}
\toprule
\multirow{2}{*}{\textbf{Train env.}} 
& \multicolumn{6}{c}{\textbf{Multimodal evaluation environment Acc. (\%)}} \\
\cmidrule(lr){2-7}
& addition\_table & mini\_sudoku & rectangle\_count & rotate\_matrix & spiral\_matrix & tictactoe\_qa \\
\midrule
Base Model
& 7.8 & 41.7 & 6.5 & 25.0 & 18.8 & 32.8 \\
\midrule
addition\_table 
& \textbf{27.8} \poschg{20.0} & 49.2 \poschg{7.5} & 17.0 \poschg{10.5} & 26.0 \poschg{1.0} & 20.0 \poschg{1.2} & 34.0 \poschg{1.2} \\
mini\_sudoku   
& 12.5 \poschg{4.7} & \textbf{72.9} \poschg{31.2} & 16.7 \poschg{10.2} & 27.1 \poschg{2.1} & 21.1 \poschg{2.3} & 35.1 \poschg{2.3} \\
rectangle\_count 
& 6.5 \negchg{1.3} & 40.0 \negchg{1.7} & \textbf{56.0} \poschg{49.5} & 24.5 \negchg{0.5} & 19.2 \poschg{0.4} & 33.2 \poschg{0.4} \\
rotate\_matrix 
& 9.3 \poschg{1.5} & 43.5 \poschg{1.8} & 19.5 \poschg{13.0} & \textbf{45.0} \poschg{20.0} & 45.8 \poschg{27.0} & 31.7 \negchg{1.1} \\
spiral\_matrix 
& 9.5 \poschg{1.7} & 43.0 \poschg{1.3} & 20.5 \poschg{14.0} & 47.3 \poschg{22.3} & \textbf{43.5} \poschg{24.7} & 31.0 \negchg{1.8} \\
tictactoe\_qa  
& 8.1 \poschg{0.3} & 41.0 \negchg{0.7} & 5.3 \negchg{1.2} & 22.1 \negchg{2.9} & 17.0 \negchg{1.8} & \textbf{45.1} \poschg{12.3} \\
\midrule
mixed6         
& 26.1 \poschg{18.3} & 59.1 \poschg{17.4} & 51.5 \poschg{45.0} & 42.5 \poschg{17.5} & 40.8 \poschg{22.0} & 39.3 \poschg{6.5} \\
$\Delta$      
& -1.7 & -13.8 & -4.5 & -2.5 & -2.7 & -5.8 \\
\bottomrule
\end{tabular}
\caption{
Transfer matrix in the multimodal setting.
}
\label{tab:mm_transfer_matrix}
\end{table*}

\subsection{Cross-Architecture Evaluation}
\label{app:internvl}

To examine whether AES is overly dependent on the model used for
annotation, we train InternVL3-8B using the environment
subset selected with Qwen3-VL-4B in Section~\ref{sec:main_results}. We do not rerun trajectory
collection, ability annotation, or environment selection for
InternVL3-8B. As shown in Table~\ref{tab:internvl}, AES substantially outperforms
Random-$K$, improving OOD-ST from 13.4 to 18.3 and OOD-MT from 8.4
to 11.8. Under HDC, the AES subset also outperforms Random-$K$ by
3.1 points on OOD-ST and 4.6 points on OOD-MT. These results provide
evidence that the selected environment distribution can transfer
across model families.

\begin{table}[t]
    \centering
    \small
    \setlength{\tabcolsep}{6pt}
    \begin{tabular}{lcc}
        \toprule
        Method & OOD-ST & OOD-MT \\
        \midrule
        Base Model          & 9.1  & 6.5  \\
        Random-$K$          & 13.4 & 8.4  \\
        AES                 & 18.3 & 11.8 \\
        Random-$K$ + HDC    & 15.8 & 10.1 \\
        \textbf{AES + HDC}  & \textbf{18.9} & \textbf{14.7} \\
        \bottomrule
    \end{tabular}
    \caption{
    Main results on InternVL3-8B. The AES
    subset is constructed using Qwen3-VL-4B.
    }
    \label{tab:internvl}
\end{table}

\subsection{More Results on AES}
\label{appendix:more_results_on_diversity}
\begin{figure*}[t]
  \includegraphics[width=\linewidth]{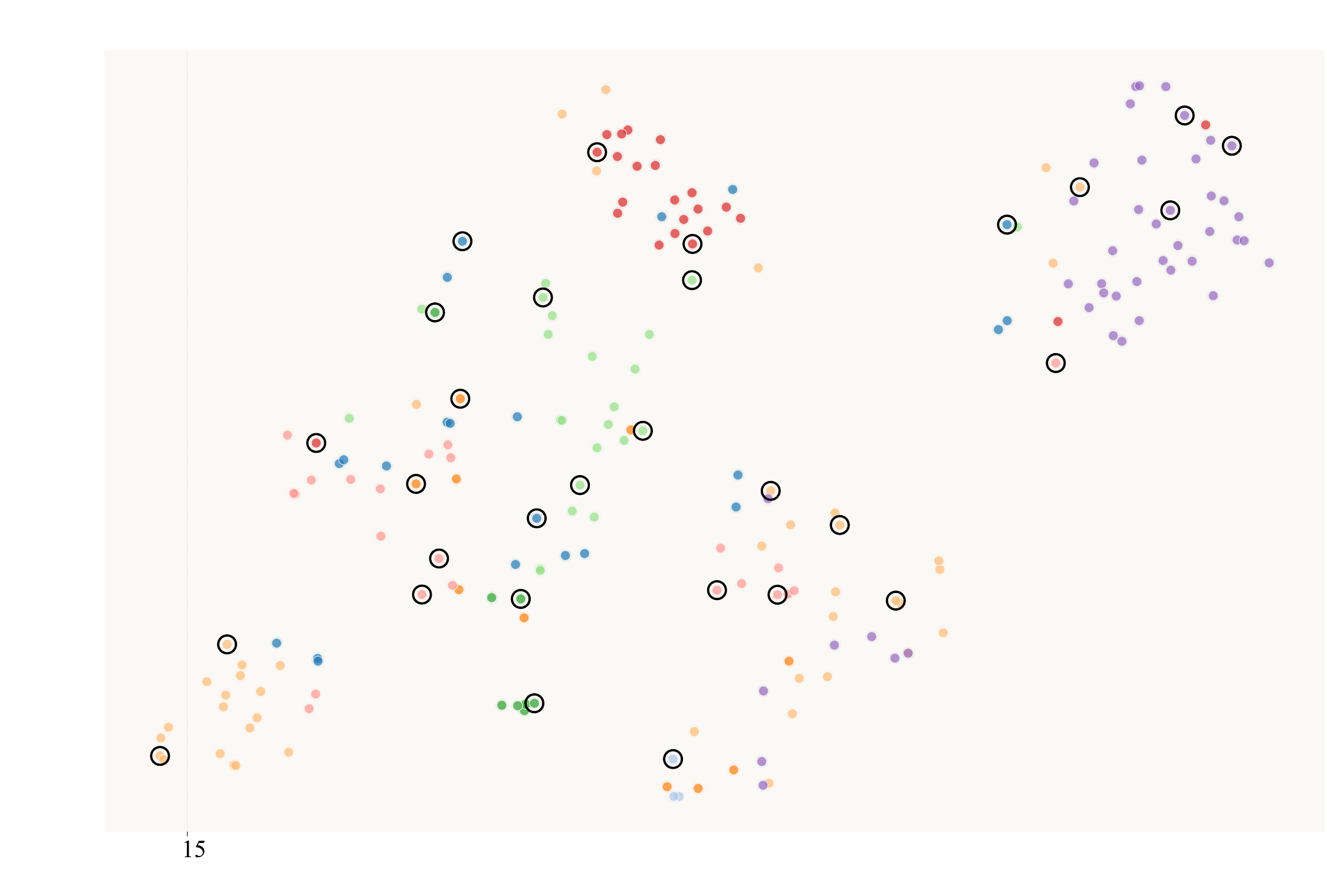}
  \caption{The distribution of the 30 environments selected by AES across the profile space of 200 environments.}
  \label{fig:selected_30_envs_tsne_appendix}
\end{figure*}

This section provides additional results for Section~\ref{sec:bcrc}. Figure~\ref{fig:selected_30_envs_tsne_appendix} shows the distribution of the selected environments in the t-SNE space of all environment profiles. The selected environments are well dispersed and cover different regions of the environment space. Our 200-environment pool is built from existing multimodal environment construction works. Therefore, the colors in the figure denote the human-defined categories from the original sources. We observe that some regions contain points of different colors, suggesting that human-defined categories do not always align with environment similarity from the perspective of agent behavior and training dynamics.

\subsection{Comparison with Existing Diversity Selection Method}
\label{appendix:surface_selection}

In Section~\ref{sec:diversity}, we mention that prior works often analyze environment diversity through surface-level information, such as representations of environment descriptions~\cite{V-GameGym}. In contrast, we argue that environment diversity should be analyzed from the perspective of the model's meta-abilities, which better reflect the underlying mechanisms required to solve the environments.

In this section, we compare AES with representation-based environment selection methods. Specifically, we consider two settings: one based on embeddings of textual environment descriptions, such as task descriptions and prompts, and the other based on embeddings of key environment code. For both settings, we use BAAI/bge-m3~\cite{bge-m3} as the embedding model and apply a cosine-distance-based greedy $k$-center selection strategy over the candidate environment pool. We start from the environment closest to the global center, and iteratively add the environment that is farthest from the current selected set. Finally, each method selects 30 environments, which are compared with the 30 environments selected by AES. We train Qwen3-VL-4B on each environment subset. As shown in Table~\ref{tab:surface_selection}, description-embedding selection brings only limited improvement, achieving a relative gain of 2.8\%. Code-embedding selection performs better, with a relative gain of 19.5\%. However, both methods are clearly worse than AES, which achieves a relative gain of 40.3\%. These results suggest that surface representations are insufficient to capture the abilities required by multimodal agents. In contrast, AES selects environments from the model's meta-ability perspective and constructs a more effective training environment distribution.

\subsection{Comparison with Curriculum-Learning Baselines.}
\label{appendix:curriculum_learning baselines}
To isolate the contribution of HDC from general curriculum learning,
we compare it with two recent curriculum-learning methods, RLVE
\citep{RLVE} and VCRL \citep{Vcrl}. All methods use
the same AES-selected environment subset and the same total training
budget. As shown in Table~\ref{tab:curriculum_baselines}, HDC achieves
the best performance across all evaluation settings. The improvements are particularly pronounced on
multi-turn environments: HDC outperforms the strongest competing
baseline by 5.9 points on ID-MT and 3.1 points on OOD-MT. These results
demonstrate that harness weakening provides additional benefits beyond
conventional curriculum learning.

\begin{table}[t]
    \centering
    \small
    \setlength{\tabcolsep}{4pt}
    \begin{tabular}{lcccc}
        \toprule
        Method & ID-ST & ID-MT & OOD-ST & OOD-MT \\
        \midrule
        AES                 & 37.7 & 25.4 & 21.0 & 12.2 \\
        AES + RLVE          & 39.9 & 28.2 & 21.2 & 13.7 \\
        AES + VCRL          & 42.5 & 30.3 & 21.6 & 12.8 \\
        \textbf{AES + HDC}  & \textbf{45.0} & \textbf{36.2}
                            & \textbf{22.1} & \textbf{16.8} \\
        \bottomrule
    \end{tabular}
    \caption{
    Comparison with curriculum-learning baselines. All methods use
    the same AES-selected environments and matched training budget.
    ST denotes single-turn success rate and MT denotes normalized
    return in multi-turn environments.
    }
    \label{tab:curriculum_baselines}
\end{table}

\subsection{Training Curves of HDC}
\label{appendix:training_curve_of_HDC}

\begin{figure*}[t]
  \includegraphics[width=\linewidth]{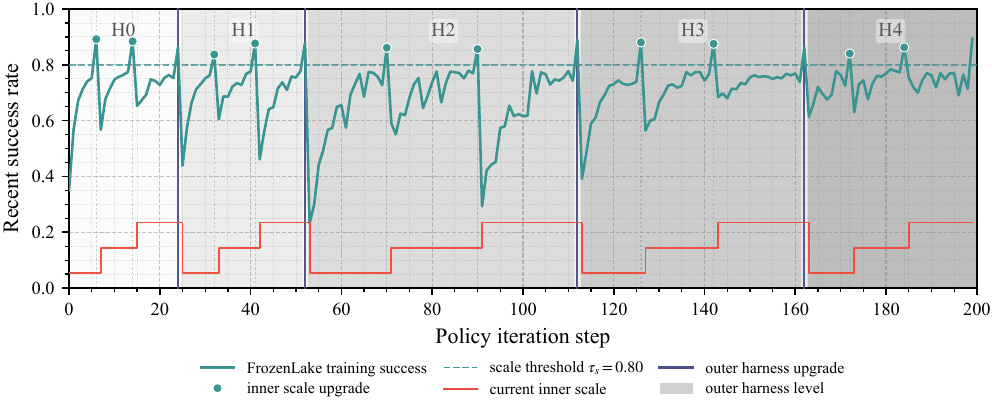}
  \caption{The training curve of HDC on Frozen Lake.}
  \label{fig:training_curve_HDC}
\end{figure*}

This section further illustrates the effectiveness of our bi-level curriculum. We train a model on the Frozen Lake environment alone and visualize the recent success rate as training progresses in Figure \ref{fig:training_curve_HDC}. The shaded regions represent different harness levels, corresponding to different stages of the outer curriculum. The curve shows an adaptive learning process. At a fixed difficulty level, the model performance gradually improves. Once the reward reaches the threshold, the curriculum advances to the next state-scale level. After the target scale is completed, the harness level is reset to a harder setting, and the state-scale progression starts again.

\subsection{Error Correction Analysis after HDC}
\label{appendix:error_analysis}
\begin{figure*}[t]
  \includegraphics[width=\linewidth]{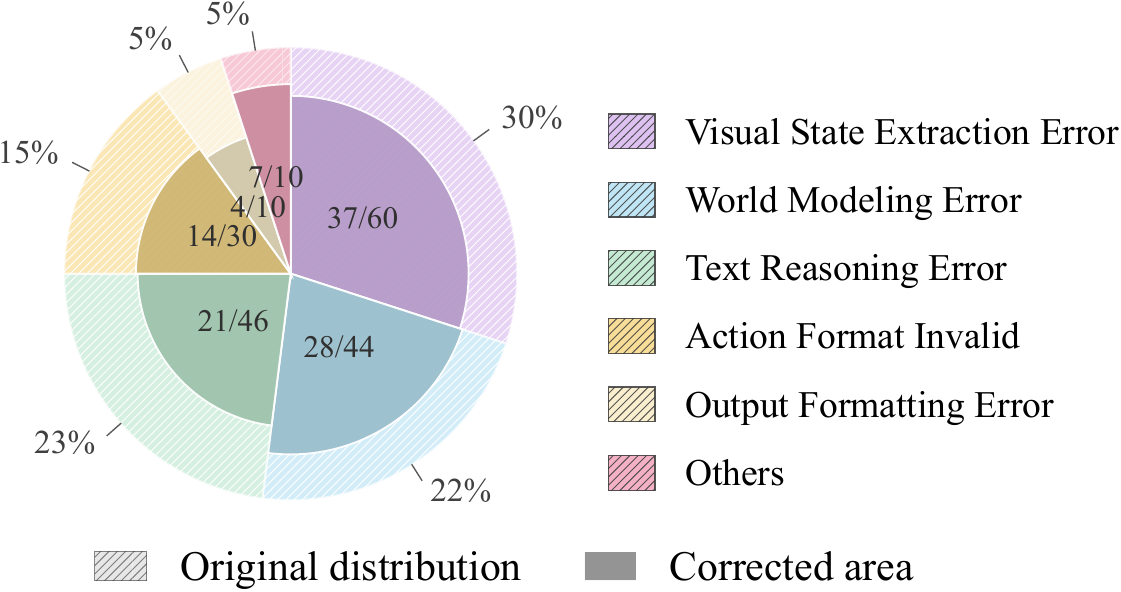}
  \caption{Corrected error analysis when learning algorithm using HDC.}
  \label{fig:corrected_error_analysis}
    \vspace{-10pt}
\end{figure*}

To further analyze how HDC changes the model's ability, we evaluate the trained model on the 200 samples used in our previous error analysis. We focus on whether the model can correct the previously observed failure modes after training. As shown in Figure~\ref{fig:corrected_error_analysis}, HDC substantially reduces errors related to visual state extraction and world modeling, which are exactly the two major bottlenecks targeted by our curriculum design. These results provide additional evidence that HDC does not merely improve the final success rate, but also directly mitigates the key multimodal reasoning failures identified in our analysis.

\textbf{Clarification on Human Annotation.}
The analysis in this section and Section~\ref{sec:bottleneck} was conducted manually by human annotators. The annotators were members of the research team or internal collaborators. Before annotation, they received the necessary disclosures and provided consent, and the work was supported through institutional funding. We provided clear guidelines, including the definition of each error category and representative examples. For ambiguous cases, the final labels were determined through internal discussion.

\section{Prompt for Atomic Ability Annotation}
\label{appendix:atomic_ability_prompt}

This appendix presents the prompt used to annotate model rollout trajectories into atomic behavioral steps and atomic abilities. The annotations are used to construct environment-level ability profiles and ability graphs.


\vspace*{\fill}

\begin{table*}[h]
\caption{
System prompt for atomic behavioral step and atomic ability annotation.
}
\label{appendix:atomic_ability_system_prompt}
\centering
\small
\begin{tabular}{p{0.97\linewidth}}
\toprule
\underline{\textbf{Prompt for Atomic Ability Annotation}} \\
\vspace{-1mm}

\hl{\textbf{\textsc{System Prompt}:}} \\[-1mm]

\begin{minipage}{0.97\linewidth}
\raggedright
\setlength{\parskip}{3pt}
\setlength{\parindent}{0pt}

You are an expert analyst for multimodal agent environment rollouts.

Your task is to segment a model rollout episode into reusable atomic behavioral
steps and infer solver-pattern-level atomic abilities. The annotations will be
used to:
\begin{enumerate}
    \setlength{\itemsep}{1pt}
    \setlength{\parsep}{0pt}
    \setlength{\topsep}{2pt}
    \item cluster common atomic steps across environments;
    \item build an environment-level skill graph;
    \item measure ability overlap and redundancy between environments;
    \item support diverse environment selection.
\end{enumerate}

A rollout step is one interaction with the environment. However, an atomic
behavioral step is not necessarily identical to one environment interaction.
One atomic behavioral step may span multiple environment interactions, and one
environment interaction may also involve multiple atomic behavioral steps.
Segment the trajectory according to the model's functional problem-solving
process, rather than the simulator's step granularity.

Use solver-pattern-level labels. Each label should be abstract enough to transfer
across environments, but concrete enough to distinguish different solution
mechanisms.

Good labels include:
\texttt{explore\_unseen\_area},
\texttt{extract\_grid\_or\_matrix\_state},
\texttt{extract\_graph\_or\_relation\_state},
\texttt{solve\_index\_mapping\_or\_traversal},
\texttt{manipulate\_object},
\texttt{solve\_constraint\_satisfaction},
\texttt{solve\_grid\_propagation},
\texttt{simulate\_rule\_based\_dynamics},
\texttt{generate\_legal\_actions},
\texttt{update\_belief\_from\_feedback},
\texttt{infer\_visual\_transformation\_rule},
\texttt{solve\_combinatorial\_counting},
\texttt{solve\_graph\_matching},
\texttt{solve\_state\_transformation}, and
\texttt{verify\_goal\_or\_terminal\_state}.

Do not use labels that are too coarse, such as
\texttt{perception}, \texttt{reasoning}, \texttt{planning},
\texttt{understanding}, \texttt{decision\_making}, or \texttt{memory}.

Do not use labels that are too fine-grained or environment-specific, such as
\texttt{compute\_first\_player\_value},
\texttt{infer\_cellular\_automaton\_rule},
\texttt{read\_arrow\_at\_current\_cell},
\texttt{identify\_sudoku\_empty\_cell},
\texttt{map\_cell\_coordinates\_under\_rotation},
\texttt{maintain\_spiral\_boundary\_indices},
\texttt{count\_live\_neighbors\_in\_game\_of\_life},
\texttt{pick\_Alice\_move}, or
\texttt{move\_to\_red\_box\_in\_TMaze}.

Instead, map environment-specific behaviors to reusable solver-pattern labels.
For example, \texttt{compute\_first\_player\_value} should be mapped to
\texttt{solve\_adversarial\_search};
\texttt{infer\_cellular\_automaton\_rule} should be mapped to
\texttt{simulate\_rule\_based\_dynamics};
\texttt{read\_arrow\_at\_current\_cell} should be mapped to
\texttt{model\_transition\_or\_operation\_rule};
\texttt{identify\_sudoku\_empty\_cell} should be mapped to
\texttt{solve\_constraint\_satisfaction};
\texttt{map\_cell\_coordinates\_under\_rotation} and
\texttt{maintain\_spiral\_boundary\_indices} should be mapped to
\texttt{solve\_index\_mapping\_or\_traversal};
and \texttt{move\_to\_red\_box\_in\_TMaze} should be mapped to
\texttt{solve\_path\_planning} or \texttt{explore\_unseen\_area}.

For Sudoku-like constraint puzzles, use
\texttt{verify\_solution\_constraints} for both partial legality checks and final
solution validation. Do not emit a separate
\texttt{detect\_inconsistency} label for Sudoku.

An atomic behavioral step is a locally coherent functional segment in the rollout,
such as extracting or updating structured state, building or updating a spatial
map, modeling a rule, constraint, transition, or legal action set, choosing or
executing a path-planning subgoal, satisfying an action precondition,
manipulating an object, verifying success, failure, legality, or terminal state,
or recovering from a wrong path or failed assumption.

Follow these segmentation rules:
\begin{enumerate}
    \setlength{\itemsep}{1pt}
    \setlength{\parsep}{0pt}
    \setlength{\topsep}{2pt}
    \item merge consecutive environment interactions if they serve the same immediate function;
    \item split when the functional role changes;
    \item split at key state-changing or informative events, such as pickup, drop, toggle, unlock, collision, death, goal reached, invalid action, feedback received, backtrack, or retry;
    \item for failed trajectories, still segment by attempted functions;
    \item prefer stable reusable labels over environment-specific labels;
    \item use the same label for similar solver patterns across categories.
\end{enumerate}

Return exactly one valid JSON object only. Do not wrap the JSON in markdown fences.
Each atomic step must contain \texttt{start\_index} and \texttt{end\_index}
referring to rollout indices. Each atomic step must include evidence from the
rollout. Include failure modes for failed or mixed rollouts. Include an
\texttt{environment\_skill\_graph} with nodes and directed edges. Also include
\texttt{cross\_category\_skill\_tags} for downstream clustering. Do not include
any explanations outside the JSON object.

\end{minipage}
\\
\bottomrule
\end{tabular}
\end{table*}

\vspace*{\fill}

\appendix

\end{document}